\documentclass{article}

\PassOptionsToPackage{numbers, square}{natbib}

 \usepackage[preprint]{neurips_2026}

\usepackage[utf8]{inputenc} 
\usepackage[T1]{fontenc}    
\usepackage{hyperref}       
\usepackage{url}            
\usepackage{booktabs}       
\usepackage{amsfonts}       
\usepackage{nicefrac}       
\usepackage{microtype}      
\usepackage{xcolor}         
\usepackage{graphicx} %
\usepackage{enumitem} 
\usepackage{url} 
\usepackage{algorithm} 
\usepackage{algpseudocode} 
\usepackage{amsmath} 
\usepackage{arydshln} 

\title{AI Outperforms Humans in Personalized Image \\ Aesthetics Assessment via LLM-Based Interviews \\ and Semantic Feature Extraction}

\author{%
  Yoshia Abe ~~~~~~~~~~Tatsuya Daikoku ~~~~~~~~~~Yasuo Kuniyoshi\\
  Graduate School of Information Science and Technology\\
  The University of Tokyo\\
  Bunkyo-ku, Tokyo 113-8654, Japan\\
  \texttt{\{y-abe, daikoku, kuniyosh\}@isi.imi.i.u-tokyo.ac.jp} \\
}

\begin{document}

\maketitle

\begin{abstract}

Accurately predicting individual aesthetic evaluation for images is a fundamental challenge for AI. 
Various deep learning (DL)-based models have been proposed for this task, training on image evaluation data to extract objective low-level features. 
However, aesthetic preferences are inherently subjective and individual-dependent. 
Accurate prediction thus requires the extraction of high-level semantic features of images and the active collection of preference information from the target individual. 
To address this issue, we focus on the utility of Large Language Models (LLMs) pretrained on vast amounts of textual data, and develop an integrated DL-LLM system.
The system actively elicits aesthetic preferences through LLM-based semi-structured interviews and predicts aesthetic evaluation by leveraging both low-level and high-level features. 
In our experiments, we compare the proposed system against conventional systems, human predictors, and the target individual's own re-evaluations after a certain time interval.
Our results show that the proposed system outperforms all of them, with particularly strong performance on highly-rated images. 
Moreover, the prediction error of the proposed system is smaller than within-person variability, while human predictors show the largest error, likely due to the influence of their own aesthetic values.
These results suggest that AI may be better positioned than others or one's future self to capture individual aesthetic preferences at a given point.
This opens a new question of whether AI could serve as a deeper interpreter of human aesthetic sensibility than humans themselves. 
\end{abstract}

\section{Introduction}

How accurately can AI predict your aesthetic preferences?
Research on AI-based aesthetic assessment has evolved from machine learning (ML) to deep learning (DL) techniques, focusing primarily on images as the target modality \citep{valenzise2022advances}. 
Early work addressed the Generic Image Aesthetics Assessment (GIAA) task, which aims to predict the average aesthetic evaluation of a population \citep{li2020personality}. 
However, it has become increasingly evident that aesthetic experience varies across individuals, and that personal aesthetic evaluation cannot be adequately described by population averages alone \citep{ren2017personalized, zhu2022personalized}. 
Motivated by this limitation, recent attention has shifted toward the Personalized Image Aesthetics Assessment (PIAA) task, which seeks to predict the aesthetic evaluations of specific individuals \citep{li2020personality}. 
Over the past several years, a variety of methods have been proposed to address the fundamental challenge of PIAA---namely, the scarcity of per-individual rating data \citep{ren2017personalized, park2017personalized, cui2020personalized}. 
However, individual aesthetic evaluation is shaped by subjective and internal information processing \citep{reber2004processing}, suggesting that conventional approaches that rely solely on objective, physical low-level features are inherently limited. 
To adequately capture personal aesthetic preferences, it is necessary to leverage high-level features that encode the semantic content and contextual meaning of images \citep{he2022rethinking}; moreover, effectively eliciting such high-level features at the individual level is well facilitated through linguistic interaction with the target person. 
To this end, we argue that large language models (LLMs)---trained on vast corpora of text and endowed with human-like prior knowledge \citep{hollman2023large}---offer an effective means of enabling such interaction. 

In this work, we develop an integrated DL-LLM system that (1) linguistically captures the aesthetic preferences of a target individual through an LLM-based semi-structured interview, and (2) predicts aesthetic evaluation scores from both low-level and high-level feature perspectives based on the elicited personal preferences. 
We evaluate the proposed system through a multi-session user study, benchmarking its prediction accuracy against not only conventional models, but also human predictors and the participants' own re-evaluations after a certain time interval. 
This experimental design enables us to make both an engineering contribution---demonstrating performance gains over conventional PIAA methods---and a broader scientific contribution, by quantitatively characterizing aesthetic evaluation behaviors across three parties: AI, other humans, and the individual themselves. 

The contributions of this work are as follows: 
\begin{enumerate}[
  before=\vspace{0em}, 
  after=\vspace{0em},    
  labelwidth=0.5em, 
  labelsep=0.5em, 
  align=left,      
  leftmargin=2em  
  ]
    \item We develop an LLM-based interview system that actively elicits the aesthetic preferences of a target individual, along with a prediction system integrating DL models and LLMs that automatically extracts semantic features to construct a personalized prediction model.
    \item Through a detailed user study, we compare the prediction performance of the proposed system against a diverse set of predictors (DL models, LLMs, other humans, and the individuals themselves). We demonstrate that our system outperforms all baselines and achieves particularly high accuracy for highly-rated images.
    \item We additionally report on the temporal variability of individual evaluation and the inherent difficulty of interpersonal prediction, offering new perspectives on the uniqueness of human aesthetic sensibility and the positioning of AI in the alignment to human aesthetic preferences. 
\end{enumerate}

\section{Related works}

Efforts to computationally approximate human aesthetic evaluation of visual stimuli have been studied in the field of computational aesthetics \citep{hoenig2005defining}.
Early approaches built predictive models using handcrafted or generic image features with classical machine learning methods; with the advent of deep learning, models leveraging deep features extracted by neural networks subsequently emerged \citep{valenzise2022advances}. 
Since training deep learning models requires large amounts of data, large-scale datasets were constructed \citep{murray2012ava}, and the field primarily addressed the Generic Image Aesthetics Assessment (GIAA) task, which aims to predict the mean aesthetic rating assigned to each image by multiple annotators \citep{li2020personality}.
However, because human aesthetic evaluation is strongly individual-dependent \citep{reber2004processing, manovich2024artificial}, relying on population averages risks discarding information about the characteristic behavior of each individual \citep{ren2017personalized, zhu2022personalized}.
Inspired by this insight, the Personalized Image Aesthetics Assessment (PIAA) task---which takes the aesthetic ratings of specific individuals as prediction targets---began to receive attention from the late 2010s onward \citep{li2020personality}. 
To address the difficulty of collecting large-scale rating data on a per-individual basis, early PIAA research saw the development of architectures that disentangle the learning of population-level average prediction from individual-level rating prediction \citep{ren2017personalized, park2017personalized}, as well as methods that repurpose other behavioral data as proxies for aesthetic ratings \citep{cui2020personalized}. 
Nevertheless, directly predicting aesthetic evaluations---which depend on subjective internal processing---from objective image features alone has inherent limitations. 
Motivated this recognition, some methods have been proposed that incorporate individual-level attribute information \citep{yang2022personalized}, or augment the pipeline with a module that infers personal characteristics from images \citep{li2020personality}.

More recently, attempts have been made to leverage the prior knowledge embedded in large pretrained models. 
One such approach couples an image processing module with a pretrained LLM to perform aesthetics assessment using user profile information \citep{wang2026enhancingzeroshotpersonalizedimage}.
Another study proposes an LLM-based PIAA prediction method and demonstrates that focusing on high-level features is effective for improving prediction performance, particularly for highly-rated images \citep{abe2025harnessingthepowerofllms}.
Following the study \citep{abe2025harnessingthepowerofllms} and prior studies \citep{deng2017image, peters2007aesthetic}, we here define low-level and high-level image features as follows. 
Low-level image features are those derived from the physical attributes of an image and can be extracted without semantic understanding; high-level image features are those extracted through semantic understanding of the subject matter and context contained in the image. 
The former includes hue, saturation, brightness, contract, edges, textures, and geometric structure, while the latter encompasses subject matter, narrative, emotion- and body-related cues, and cultural, social, and historical information.

Meanwhile, in the field of automated machine learning---distinct from aesthetics assessment---research has emerged on methods that leverage foundation models pretrained on large-scale data to automatically extract high-level features. 
For example, one approach applies CLIP \citep{radford2021CLIP}---a vision-language model trained to align images and text---to extract and name high-level concepts from images for use in classification tasks \citep{rao2024DNCBM}. 
Another prompts an LLM to devise new features for tabular data using domain knowledge, iteratively improving the predictive performance of a machine learning model \citep{hollman2023large}. 
These approaches suggest that automatically extracting and linguistically describing high-level features can effectively support predictive modeling. 
Aesthetics assessment is a domain where such semantic and subjective factors play an important role, making it a natural fit for this paradigm.
Yet it remains unexplored in this context, and the present work addresses this gap.

There is another advantage to using LLMs for aesthetic assessment prediction tasks: they enable linguistic communication with the target individual. 
Since aesthetic evaluations are strongly influenced by a person's past experiences and unique values, directly eliciting that individual's aesthetic preferences and values through language is considered a particularly direct and effective approach. 
There are some precedents in adjacent fields---such as personal modeling research \citep{park2024generativeagentsimulations1000} and recommendation systems \citep{yun2025userexperience}---for extracting a person's preferences, values, or beliefs through linguistic interaction. 
In addition, although designed for the GIAA task, a method exists that learns from linguistic annotations attached to images to perform aesthetic assessment \citep{ke2023vila}.
Nevertheless, the idea of using verbal data about aesthetic preferences obtained through interaction with the target individual has never been implemented or empirically validated for the PIAA task.

\section{Methods}

In this work, we address the PIAA task by developing a system that goes beyond objective image features to capture subjective aesthetic preferences shaped by personal experiences and contextual background. 
The proposed system consists of two subsystems: an Interview System that elicits aesthetic preferences through an LLM-automated interview, and a Prediction System that extracts high-level features from linguistic and image rating data to train a personalized predictive model.

\subsection{Data collection of individual aesthetic ratings}

We conducted a participant study involving multiple behavioral measurements and interviews to collect data on individual aesthetic preferences and to evaluate the performance of the trained system. 
A total of 30 individuals (all in their 20s; 25 male, 5 female) were recruited online, and informed consent was obtained from all participants.
Further details are provided in Appendix~\ref{App-subsec:participant_study}.

In the study, participants rated photographic images randomly sampled from the PARA dataset~\citep{yang2022personalized} on an aesthetic scale from 1.0 to 5.0 (increments of 0.5). 
Five semantic categories were used, inspired by prior work~\citep{nieto2022understanding}: portrait, animal, scene, building, and plant.
Images were divided into three rating classes  (Low, Middle, and High) based on mean annotator scores from the dataset, and 20 images were randomly sampled from each category-class combination, yielding \(n_{\mathrm{image}} = 300\) image samples---informed by prior studies using approximately 200 images per rater~\citep{ren2017personalized, kong2016photo} and recommendations suggesting over 100 images for individual prediction~\citep{abe2025quantitative}.

\subsection{Capturing aesthetic preferences via LLM-based interview}
\begin{figure}[htbp]
  \centering
  \includegraphics[width=0.8\linewidth, trim=0 740 1365 0, clip]{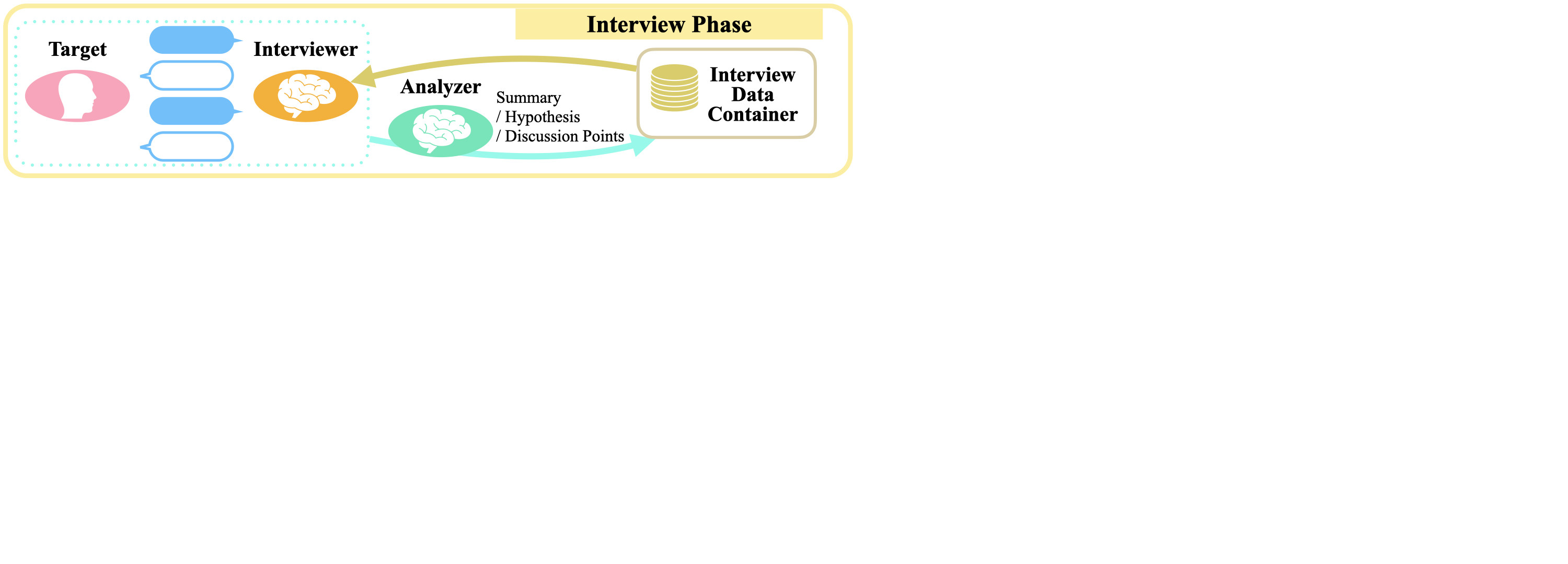}
  \caption[Overview of the interview system]{Overview of the Interview System. 
The system consists of two asynchronous agents (the Interviewer agent and the Analyzer agent) operating in parallel. 
Both agents observe a shared interview data container that stores the interview information.
}
  \label{fig:interview_system}
\end{figure}

The Interview System has the structure shown in Fig.~\ref{fig:interview_system}.
By conducting a semi-structured interview rather than a fixed-question interview, the system can flexibly formulate questions based on the target individual's responses, enabling deeper and more context-sensitive understanding. 
The system is a multi-agent system in which two agents operate asynchronously in parallel: an Interviewer agent that poses questions to the target individual, and an Analyzer agent that analyzes the individual's responses. 
Both agents operate while continuously observing a shared data structure called the Interview Data Container, which stores the interview themes, the specific points to be covered, the conversation history, and the analysis results. 
Three themes are used for the interview: Preference Targets, Image-Evoked Reactions, and Personal Tastes. 
A detailed list of sub-topics within each theme is provided in Appendix~\ref{App-subsec:interview_system}. 

The Analyzer generates a summary of the information obtained from the most recent response, along with insights and hypotheses derived from it, and organizes the points to be addressed for deeper exploration of preferences and comprehensive coverage of the assigned theme. 
Drawing on these outputs, the Interviewer generates the next question in a manner that flows naturally within the conversational context. 
The Interviewer is required to balance two objectives: identifying the underlying essence common to the target individual's aesthetic preferences, and achieving comprehensive understanding of the perspectives specified by the currently assigned theme. 

\subsection{LLM-based feature analysis and iterative training of ML predictive models}
\begin{figure}[htbp]
  \centering
  \includegraphics[width=\linewidth, trim=0 205 580 0, clip]{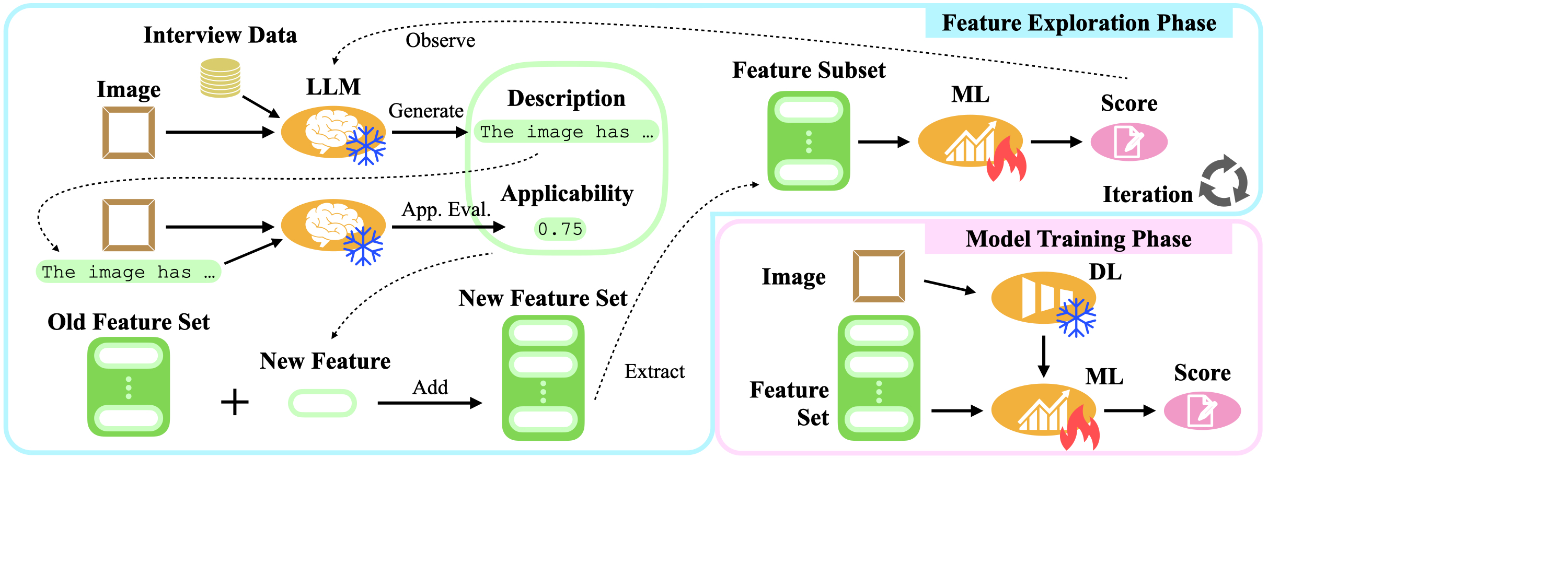}
  \includegraphics[width=\linewidth, trim=0 630 579 0, clip]{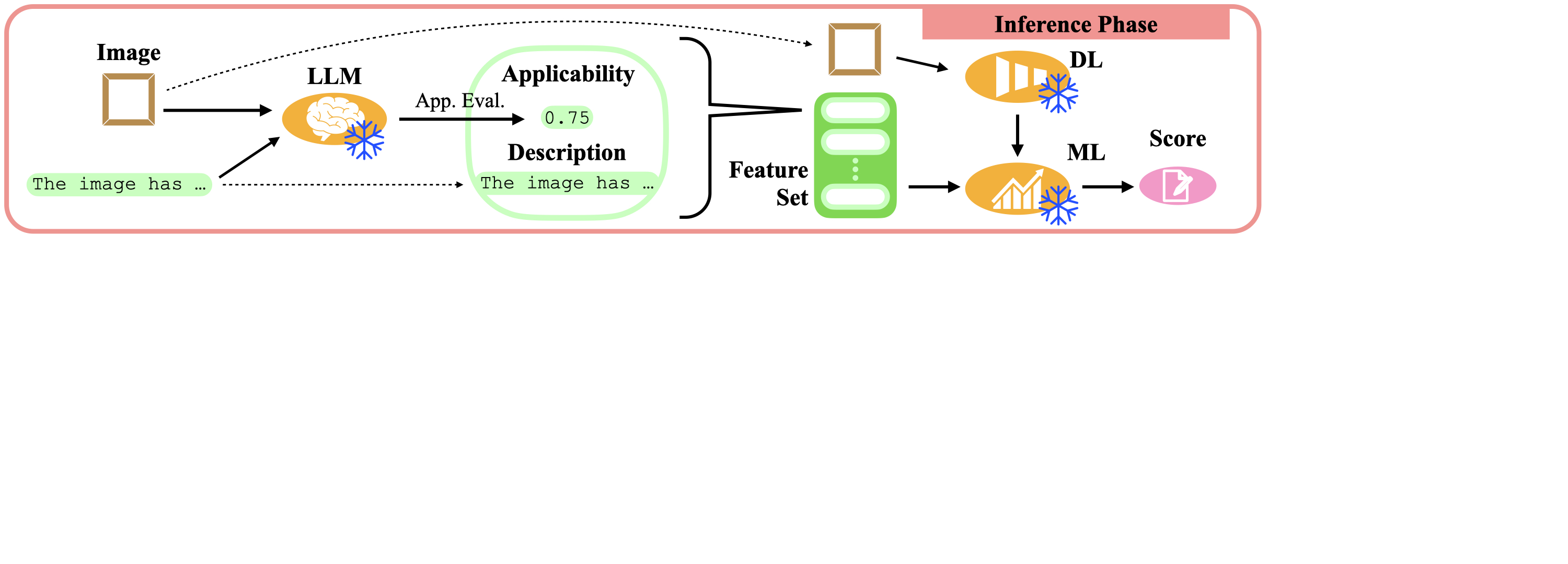}
  \caption[Overview of the Prediction system]{Overview of the Prediction System. 
The system consists of three modules: a DL module that outputs an intermediate predicted score from an image, an LLM module that generates a feature set based on the interview data and conducts applicability evaluation, and an ML module that integrates the outputs of both modules to produce the final prediction. 
The training process consists of two phases: a feature exploration phase and a model training phase. 
At inference, the system takes an image as input and outputs the predicted aesthetic score for the target individual. 
  }
  \label{fig:prediction_system}
\end{figure}

An overview of the Prediction System is shown in Fig.~\ref{fig:prediction_system}.
The system comprises a DL module that receives an image and outputs an intermediate predicted score, an LLM module that generates a feature set by extracting semantic and contextual aspects of images with reference to the interview data, and an ML module that integrates the outputs of both modules to produce the final prediction. 
The system is trained using a dataset \(D_{\mathrm{P}}\) consisting of image--rating pairs from the target individual, along with interview data, to predict the individual's rating \(p_{\mathrm{true}}\) for a given image. 
The DL module is independently pretrained on \(D_{\mathrm{P}}\) to predict \(p_{\mathrm{true}}\); accordingly, the Prediction System can also be viewed as a system that refines DL-based predictions grounded in low-level features by incorporating high-level features. 
Training of the LLM and ML modules proceeds in two phases: a feature exploration phase and a model training phase. Details of the algorithm are provided in Appendix~\ref{App-subsec:prediction_system}. 

The individual's dataset \(D_{\mathrm{P}}\) of \(n_{\mathrm{image}}\) images in total is randomly split into three sets: a training set \(D_{\mathrm{tr}}\), a validation set \(D_{\mathrm{val}}\), and a test set \(D_{\mathrm{te}}\), of sizes \(n_{\mathrm{tr}}\), \(n_{\mathrm{val}}\), and \(n_{\mathrm{te}}\), respectively. 
The training set \(D_{\mathrm{tr}}\) is further split into training and validation sets in the feature exploration phase: an inner training set \(D_{\mathrm{tr}}^{\mathrm{in}}\) and an inner validation set \(D_{\mathrm{val}}^{\mathrm{in}}\), of sizes \(n_{\mathrm{tr}}^{\mathrm{in}}\) and \(n_{\mathrm{val}}^{\mathrm{in}}\), respectively. 

The first phase, the feature exploration phase, aims to construct an appropriate feature set 
that enables accurate prediction of the individual's rating  \(p_{\mathrm{true}}\) from their rating tendencies. 
Concretely, using \(D_{\mathrm{tr}}^{\mathrm{in}}\) and \(D_{\mathrm{val}}^{\mathrm{in}}\), a temporary linear regression model is built from the current feature set with \(p_{\mathrm{true}}\) as the target variable; the LLM observes the prediction errors and devises new features to add; this process is repeated multiple times. 
When devising new features, the LLM is provided with the explanatory variables and regression coefficients of the current predictive model, the existing feature set, image samples with large prediction errors, and the interview data (the Analyzer's analysis results and the Interviewer's summary comments), which serve as the basis for its reasoning. 
Here, the value of a given feature represents a numerical score assigned by the LLM indicating how well a linguistic description of a specific image property applies to that image (referred to as applicability). 
In essence, creating a new feature amounts to creating a linguistic description of a specific image property; once such a description is defined, the LLM automatically evaluates the degree to which each image exhibits that property (i.e., the applicability value). 

The second phase, the model training phase, constructs the final ML model with \(p_{\mathrm{true}}\) as the target variable, using the feature set obtained in the feature exploration phase and the intermediate predicted scores from the DL module, which is pretrained as described in Sections~\ref{subsec:configuration_of_PS} and~\ref{subsec:baselines}. 
A hyperparameter search is conducted on the training set  \(D_{\mathrm{tr}}\) and validation set  \(D_{\mathrm{val}}\), and the best-performing instance is adopted as the final predictive model. 

At inference time, the LLM evaluates the applicability of each feature in the constructed feature set for a given image; the ML module then combines these applicability values with the intermediate predicted score from the DL module to produce the final prediction. 

\section{Experimental setup}

\subsection{Evaluation protocol}
Prediction performance was evaluated on the test set \(D_{\mathrm{te}}\).
We adopted mean absolute error (MAE) as the metric for measuring PIAA task performance. 
The Spearman rank correlation coefficient, which has been used in prior work \citep{ren2017personalized, zhu2022personalized}, measures consistency in the ranking of predictions; however, it cannot quantify the magnitude of prediction error, which is particularly problematic in aesthetics assessment where score baselines and scales differ across individuals. 
Furthermore, since we directly compare the prediction error of the proposed system with within-person variability, MAE is appropriate as it allows both to be measured on the same scale. 
MAE also has the advantage of being computable per image, enabling more fine-grained analysis than Spearman rank correlation.
MAE was measured for all 30 participants, and the distribution and summary statistics were compared. 

Regarding the dataset split, the \(n_{\mathrm{image}} = 300\) images were first partitioned into the test set \(D_{\mathrm{te}}\) (\(n_{\mathrm{te}} = 45\)) and the remaining 255 images, which were then split into  
the training set \(D_{\mathrm{tr}}\) and validation set  \(D_{\mathrm{val}}\) at a ratio of \(n_{\mathrm{tr}}: n_{\mathrm{val}} = 4:1\). 
The training set was further split into 
\(D_{\mathrm{tr}}^{\mathrm{in}}\) and \(D_{\mathrm{val}}^{\mathrm{in}}\) at a ratio of \(n_{\mathrm{tr}}^{\mathrm{in}}: n_{\mathrm{val}}^{\mathrm{in}} = 3:1\).
The test set was held out and shared across all comparisons.

\subsection{Configuration of the proposed system}
\label{subsec:configuration_of_PS}
Regarding the LLMs used, both the Interviewer agent and the Analyzer agent employed 
Claude Sonnet 4.5~\citep{anthropic2025introducingclaudesonnet45}.
For the Prediction System, Claude Sonnet 4.5 was used for feature generation, 
Gemini 2.5 Flash Lite~\citep{google2025gemini25} was used for applicability evaluation, 
and Gemini 2.5 Flash~\citep{google2025gemini25} was used for retry processing upon API call errors. 
In all cases, the temperature was set to 0.0 and the maximum output token length to 4096. 
For the ML module, we adopted a gradient boosting regressor (GBR) as the predictive model, chosen for its ability to flexibly capture nonlinear relationships between features and ratings as well as feature interactions in aesthetics assessment. 
For the DL module, the DL-based predictor described in the following subsection was trained across five independent runs in advance, and the mean of their predicted scores was used as input to the ML module. 

\subsection{Baselines}
\label{subsec:baselines}
We compared the proposed system against the following baselines: a DL-based predictor, LLM-based predictors, and human predictors. 
Brief descriptions are provided as follows (see Appendix~\ref{App-subsec:predictor_configurations} for details).
The DL-based predictor was constructed following a prior study \citep{abe2025quantitative} by training a convolutional neural network with a ResNet-50 backbone~\citep{he2015deep} on the GIAA data of the PARA dataset, followed by fine-tuning on the target individual's rating data. 
To account for training randomness, the final prediction was obtained by averaging the outputs of five independently trained model instances. 
The LLM-based predictor was constructed following a prior study \citep{abe2025harnessingthepowerofllms} as a system that infers the target individual's aesthetic tendencies from 12 few-shot examples and predicts the aesthetic rating of a given test image on a scale from 1.0 to 5.0 in increments of 0.5.
The LLMs used were Claude Sonnet 4.5/3.7~\citep{anthropic2025introducingclaudesonnet45, anthropic2025claude37}, 
GPT-5/4o~\citep{openai2025gpt5, openai2024hello}, 
Gemini 2.5 Flash/2.0 Flash~\citep{google2025gemini25, google2024next}. 
For the human predictors, each of the 30 participants predicted the ratings of 5 anonymized individuals selected from the remaining 29 participants. 
Specifically, each predictor was presented with the rating data of the target individual from the training and validation sets (255 images in total) along with the interview data (the analysis results and the summary), and was asked to learn the target individual's tendencies and summarize them concisely in their own words. 
The predictor then predicted ratings for the 45 test images on a scale from 1.0 to 5.0 in increments of 0.5. 

\begin{figure}[htbp]
  \centering
  \includegraphics[width=\linewidth, trim=0 0 0 0, clip]{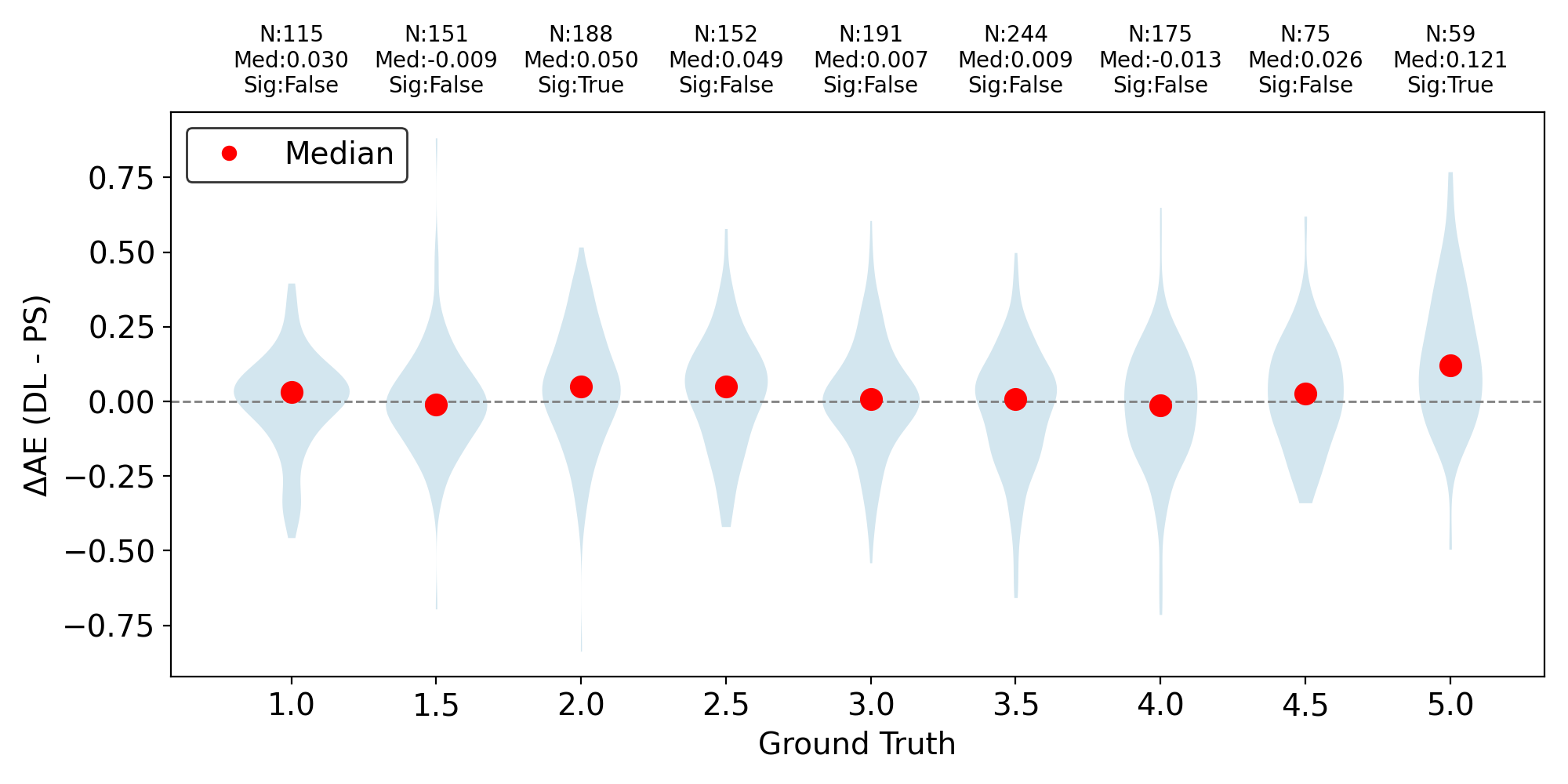}
  \caption[~]{
Violin plots of the per-sample difference in absolute prediction error between the DL-based predictor (DL) and the proposed system (PS), defined as DL minus PS, grouped by ground-truth rating values. 
Annotations at the top report sample size (N), median (Med), and statistical significance (Sig) from  Wilcoxon signed-rank tests against zero with Benjamini-Hochberg correction (significance level: 0.05).
PS shows the greatest advantage over DL in the rating-5 region, suggesting that high-level features are particularly effective for images that evoke strong aesthetic appreciation. 
  }
  \label{fig:diff_distribution_by_gt}
\end{figure}

\section{Results}

\subsection{Improvement over the DL-based predictor}
\paragraph{Overall improvement}
The proposed system is designed to refine DL-based predictions grounded in low-level features by additionally incorporating high-level features derived from the target individual's aesthetic preferences. 
We therefore first examine the improvement over the DL-based predictor. 
The mean (\(\pm\) standard deviation) of MAE across 30 participants was 0.561 (\(\pm\) 0.110) for the DL-based predictor and 0.549 (\(\pm\) 0.116) for the proposed system. 
A Wilcoxon signed-rank test confirmed that the proposed system significantly outperformed the DL-based predictor (\(p = 0.0328\)), with a mean improvement of 2.3\% across 30 participants. 
Further details and ablation studies (on model training methods, predictive model types, and interview data ablation) are provided in Appendix~\ref{App-subsec:ablation_study}. 

\paragraph{Region-wise analysis}
Next, inspired by the finding in a prior study \citep{abe2025harnessingthepowerofllms} that LLM-based prediction performance for highly-rated images improves when high-level features are emphasized, 
we formulated the following hypothesis: 
since individual differences arising from subjective preferences are particularly pronounced for highly-rated images, the proposed system 
will show greater improvement over the DL-based predictor specifically in the high-rating region. 
To test this hypothesis, the 1350 rating samples in total---obtained by having 30 participants rate 45 test images each---were stratified by rating value (from 1.0 to 5.0 in increments of 0.5). 
For each rating group, we measured the distribution of the per-sample difference, defined as the absolute prediction error of the DL-based predictor minus that of the proposed system.
A positive value indicates that the proposed system has a smaller absolute prediction error and this quantity can thus be interpreted as a measure of the advantage of the proposed system over the DL-based predictor. 
The distribution plots and the results of Wilcoxon signed-rank tests against zero (significance level \(\alpha = 0.05\)) are shown in Fig.~\ref{fig:diff_distribution_by_gt}. 
In the rating-5 region, the difference between the absolute prediction errors of the DL-based predictor and the proposed system was significantly greater than zero---that is, the proposed system exhibited significantly smaller absolute prediction error than the DL-based predictor. 
This trend is also clearly visible in the shape of the violin plots and the median values. 
Furthermore, for the 59 samples with a rating of 5, the improvement rate of the proposed system was computed as the absolute error of the DL-based predictor minus that of the proposed system, divided by that of the DL-based predictor; this yielded a mean improvement of 21\% and a median improvement of 12\%.

\begin{figure}[htbp]
  \centering
  \includegraphics[width=\linewidth, trim=0 0 0 0, clip]{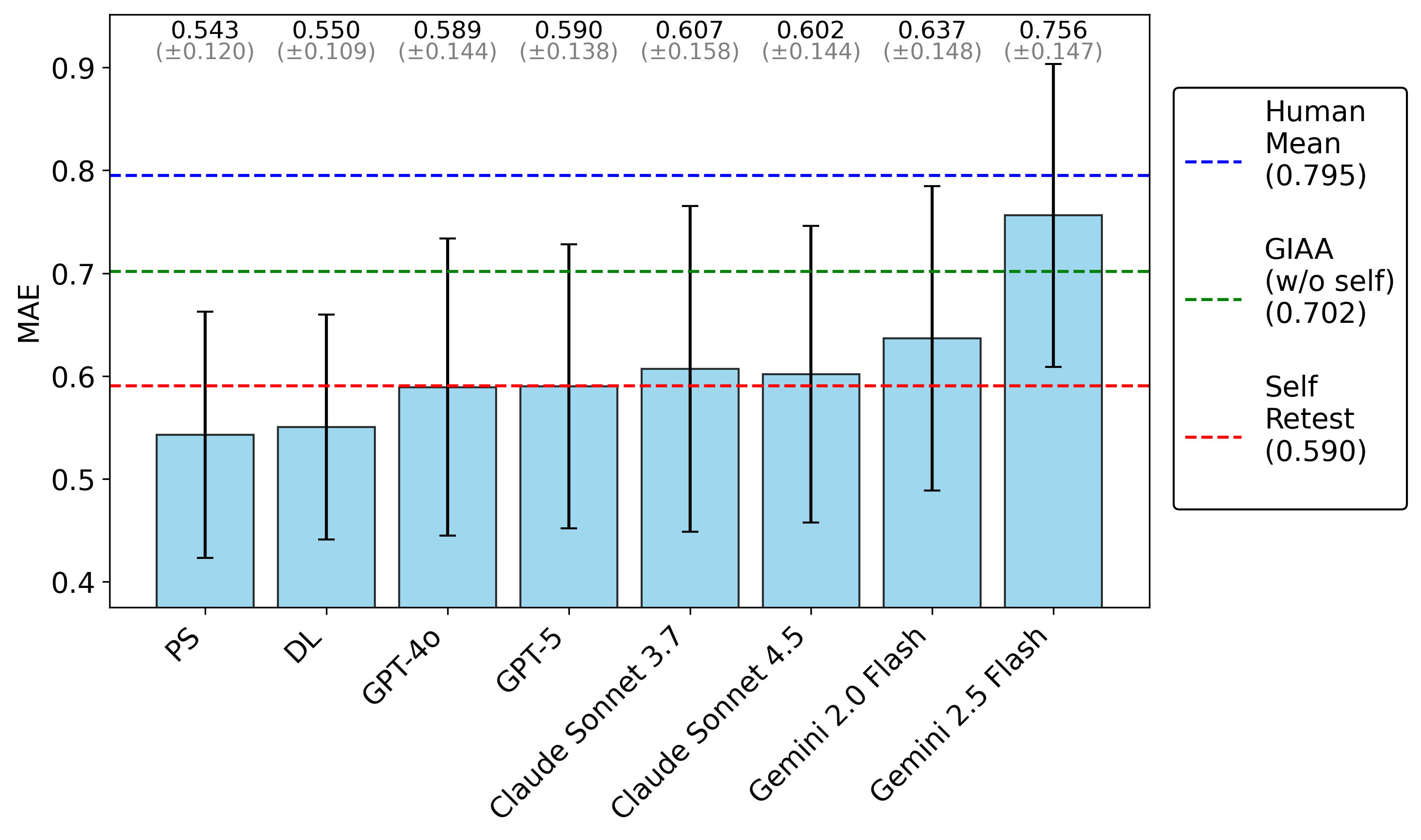}
  \caption[~]{
Comparison of MAE across all predictors. Bars show mean \(\pm\) standard deviation of MAE over 45 test images across 30 target individuals.
PS and DL denote the proposed system and the DL-based predictor, respectively; others are LLM-based predictors. 
Dashed lines indicate reference values: average MAE of human predictors (blue), population mean baseline (green), and within-person difference between original and retest ratings (red). 
The proposed system achieved the best overall performance, outperforming even retest ratings, while human predictors showed the lowest performance. 
These results suggest that AI systems can capture individual aesthetic preferences better than humans. 
}
  \label{fig:compare_all_predictors_mae_distribution}
\end{figure}

\subsection{Comparison with all predictors}
We next compared the proposed system not only against the DL-based predictor but also against all other predictors. 
Since the responses of LLM-based and human predictors take discrete values from 1.0 to 5.0 in increments of 0.5, 
the predictions of the proposed system and the DL-based predictor were clipped to the range ([1.0, 5.0]) and then rounded to the nearest multiple of 0.5 using the round-half-to-even rule, to ensure a fair comparison.

\paragraph{All predictors comparison}
The results are shown in Fig.~\ref{fig:compare_all_predictors_mae_distribution}. 
Three reference values were computed for comparison. 
Human Mean is the mean MAE of the five human predictors who predicted each target individual, averaged across all 30 target individuals. 
GIAA (w/o self) is the MAE obtained by using the mean rating of the remaining 29 participants (excluding the target individual) as the predicted value for each image, averaged across all 30 target individuals. 
Self Retest is the mean absolute difference between the original ratings and the ratings re-collected for the same 45 test images after an interval of more than two weeks, averaged across all 30 participants. 
The proposed system achieved the best performance overall, demonstrating the effectiveness of the proposed approach. 
In contrast, human predictors showed the lowest performance, indicating that it is difficult for humans to predict the aesthetic evaluations of others. 
Regarding LLM-based predictors, prediction performance varied across versions within the same model family; in particular Gemini 2.5 Flash was worse than Gemini 2.0 Flash, suggesting that technical improvements in LLMs do not necessarily benefit the task of image aesthetics assessment.
GIAA (w/o self) performed worse than most AI-based predictors, reaffirming the existence of individual differences and importance of personalization. 
Strikingly, Self Retest was outperformed by the proposed system and the DL-based predictor, demonstrating that human aesthetic evaluation is subject to substantial temporal fluctuation. 
This also suggests that an individual's aesthetic evaluation at a given moment is better captured by an AI system than by their own future self.

\begin{figure}[htbp]
  \centering
  \includegraphics[width=\linewidth, trim=0 0 0 0, clip]{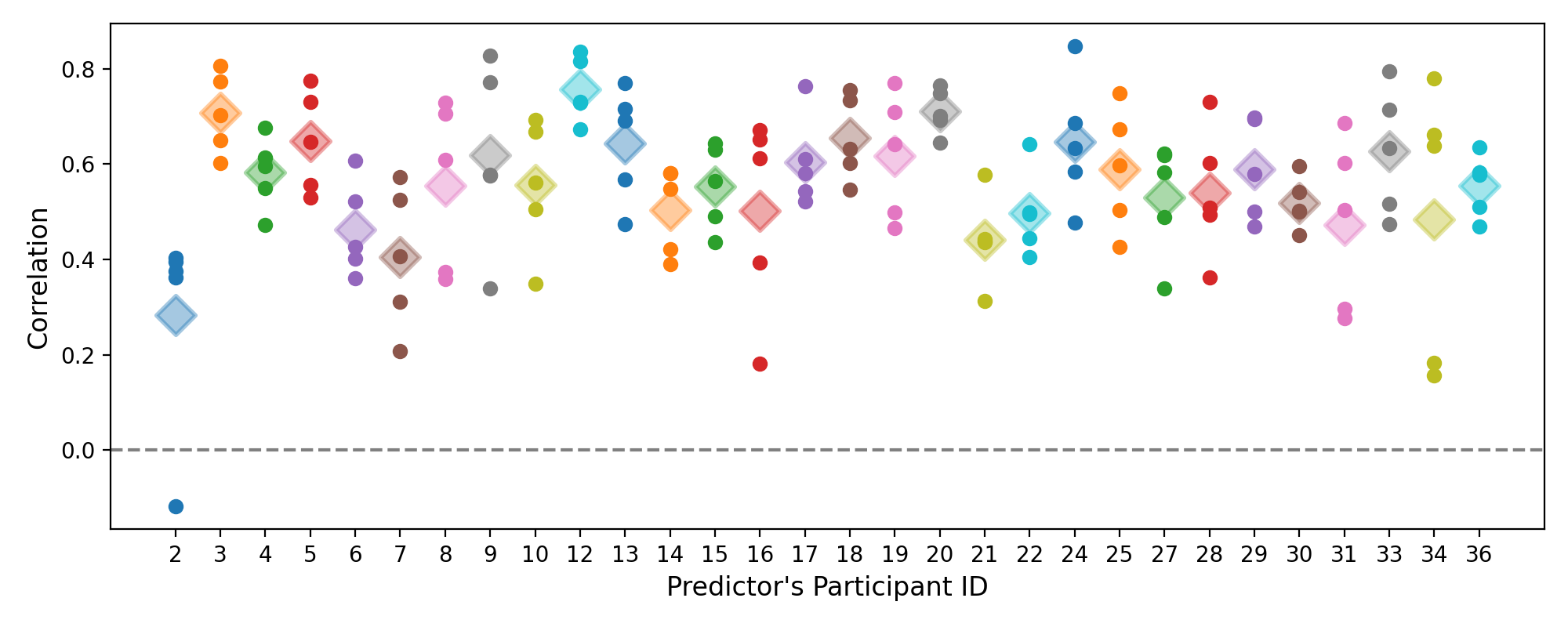}
  \caption[~]{
Pearson correlation coefficients between prediction errors \(e\) and evaluation differences \(d\) for each human predictor (N=30). The diamond markers indicate the mean across the five target individuals for each predictor.
Nearly all correlation coefficients were positive with only one exception, suggesting a systematic bias in human predictors toward their own ratings. 
  }
  \label{fig:human_predictor_correlation}
\end{figure}

\paragraph{Human prediction bias}
We also analyzed the cases in which human predictors made incorrect predictions. 
Specifically, we hypothesized that prediction errors made by human predictors tend to be biased in the direction of their own ratings, and tested this as follows. 
For a given image, let \(r_p\) denote the predictor's own rating, \(r_t\) the target individual's rating (ground truth), and \(\hat{r}_{p2t}\) the predictor's predicted rating for the target individual. 
The prediction error was defined as \(e = \hat{r}_{p2t} - r_t\), and the discrepancy between the predictor's and target individual's ratings as \(d = r_p - r_t\).
The Pearson correlation coefficient between \(e\) and \(d\) was then computed across the 45 test images. 
The results for all 30 participants acting as human predictors are shown in Fig.~\ref{fig:human_predictor_correlation}.
With only one exception showing a negative correlation, all correlation coefficients were positive, and the mean correlation coefficient averaged across the five target individuals was positive for every predictor. 
These results suggest that, as a general tendency, the direction in which a human predictor makes an error is biased toward their own rating. 
This implies that people's own value systems may act as a systematic bias that distorts their predictions of others' aesthetic evaluations---and may ultimately be what prevents them from truly understanding others' aesthetic sensibility.

\section{Discussion}

\paragraph{Main findings}
In this work, we developed an integrated DL-LLM system that linguistically captures the aesthetic preferences of a target individual through an LLM-based semi-structured interview, and predicts aesthetic ratings from both low-level and high-level feature perspectives based on the elicited personal preferences. 
Through experiments, we compared the prediction performance of the proposed system not only against conventional baselines such as DL-based and LLM-based predictors, but also against human predictors and the individuals' own future responses. 
The proposed system outperformed all predictors, with particularly pronounced gains for highly-rated images. 
This demonstrates the effectiveness of actively eliciting personal aesthetic preferences through linguistic interaction and leveraging high-level semantic features for prediction. 

\paragraph{Scientific insights}
Analysis of participant behavior further revealed two scientifically interesting insights: the temporal fluctuation of individuals' aesthetic evaluations is larger than the prediction error of AI systems, 
and human predictors' errors tend to be biased toward their own ratings.
Taken together, these findings suggest that it may be AI systems, rather than other humans or one's own future self, that can more accurately capture and predict an individual's aesthetic tendencies at a given point in time. 
The underlying mechanism may be as follows: while human aesthetic values fluctuate over time and are context-dependent, they are not flexible enough to instantly learn and emulate those of others.
AI systems, by contrast, can flexibly adapt to new individuals through fine-tuning or prompting---even if their own aesthetic values have been shaped through pretraining---and, once trained, exhibit temporal consistency without fluctuation. 

\paragraph{Broader contributions}
Beyond the main findings, the proposed system also offers broader contributions and raises directions for future work. 
Notably, participants rated their interaction experience with the Interview System positively (see Appendix~\ref{App-subsec:secondary_effects_of_IS} for details). 
The Prediction System represents a novel method that uses LLMs to extract features from unstructured image data and iteratively searches for and constructs a feature set and ML models that accurately predict the target variable. 
These constitute technical contributions with broad applicability beyond aesthetics assessment tasks. 

\paragraph{Limitations}
The current implementation has several limitations in terms of experimental scope: 
the participant pool was small (\(N=30\)) and demographically skewed (20s, predominantly male); the interviews were conducted in Japanese; the dataset consisted solely of photograph images from PARA; and the amount of data per individual was limited.
These factors may limit the applicability of the system to broader populations, cultural contexts, and image domains.
The choice of LLM and prompting strategies may also introduce model-specific biases, and the system incurs substantial computational cost due to API calls. 
More fundamentally, the difficulty of capturing aesthetic preferences that cannot be articulated linguistically, and the challenge of extracting highly variable features, may require incorporating modalities beyond language and images, such as embodied sensory information. 
Future research directions include investigating how aesthetic criteria are established and formed, and exploring how far prediction error can be reduced toward zero (\(\mathrm{MAE}=0\)).
In this sense, any gap that persists even as prediction improves may reflect the irreducible uniqueness of human aesthetic experience.

\paragraph{Broader implications}
The broader implications of modeling individual aesthetic sensibility, as pursued in this work, have both positive and negative aspects. 
On the positive side, externalizing an individual's aesthetic values as a model enables self-understanding, preservation, sharing with others, comparison, and the automation of aesthetic evaluation. 
On the negative side, it may lead to the oversimplification and fixing of values, misunderstanding of oneself and others, loss of diversity in aesthetic values through comparison and selection, and the loss of human distinctiveness as fitting AI expectations becomes prioritized, as well as potential privacy risks.
Addressing these concerns will require continued technical efforts to improve personalization accuracy and to enhance the interpretability of modeled values. A more detailed discussion is provided in Appendix~\ref{App-sec:broader_implications}.

\paragraph{Conclusion}
In summary, this work proposes a system for actively capturing and predicting individual aesthetic preferences, 
and demonstrates its effectiveness through comparative experiments with diverse predictors including humans. 
The findings provide quantitative evidence regarding the extent to which AI can capture individual aesthetic preferences, and 
raise the possibility that AI could ultimately serve as a deeper interpreter of human aesthetic sensibility than humans themselves.


\appendix

\newpage

\section{Implementation details}
The implementation details of the proposed system are described below. 

\subsection{Interview system}
\label{App-subsec:interview_system}
The interview theme categories and the list of sub-topics within each are as follows. 
Note that the originals are written in Japanese; the following are English translations for reference.  

\textbf{Preference Targets}
    \begin{enumerate}[
      before=\vspace{0em}, 
      after=\vspace{0em},    
      labelwidth=9em, 
      labelsep=0.5em, 
      align=left,      
      leftmargin=10em  
      ]
      \item [Subject] What kinds of subjects or motifs in images do you like or dislike? 
      \item [Story] What kinds of narratives or storytelling qualities in images do you like or dislike? 
      \item [Culture \& History] Do you find images that evoke a sense of cultural or historical background appealing? 
      \item [Art] What kinds of artworks, genres, or media do you typically engage with? 
      \item [Daily Moments] What kinds of scenes in everyday life make you feel beautiful, comfortable, or pleasing? Conversely, what kinds of scenes make you feel unattractive, uncomfortable, or unpleasant? 
    \end{enumerate}

  \textbf{Image-Evoked Reactions}
    \begin{enumerate}[
      before=\vspace{0em}, 
      after=\vspace{0em},    
      labelwidth=9em, 
      labelsep=0.5em, 
      align=left,      
      leftmargin=10em  
      ]
        \item [Emotional Reaction] What kinds of images evoke strong emotions or feelings in you? 
        \item [Physical Reaction] Have you ever experienced physical reactions---such as goosebumps or chills---in response to an image? 
        \item [Creativity] What kinds of images strike you as novel, original, or unique? 
    \end{enumerate}
    
  \textbf{Personal Tastes}
    \begin{enumerate}[
      before=\vspace{0em}, 
      after=\vspace{0em},    
      labelwidth=9em, 
      labelsep=0.5em, 
      align=left,      
      leftmargin=10em  
      ]
        \item [Likes] Beyond images, what kinds of things or experiences do you generally find yourself drawn to? 
        \item [Dislikes] Beyond images, what kinds of things or experiences do you generally tend to avoid or feel uncomfortable with? 
    \end{enumerate}

\subsection{Prediction system}
\label{App-subsec:prediction_system}
\begin{figure}[htbp]
  \centering
  \includegraphics[width=0.8\linewidth, trim=0 250 1200 0, clip]{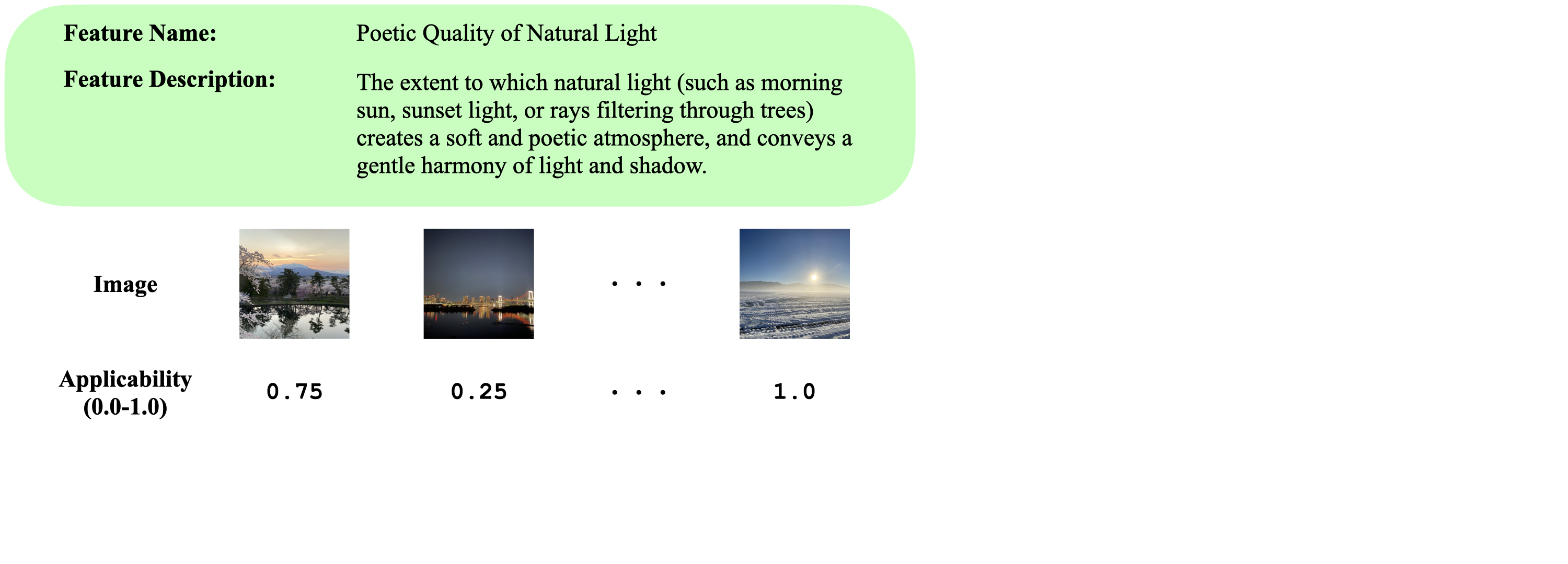}
  \caption[Example of applicability evaluation]{Example of applicability evaluation. 
          An LLM is provided with a feature name, its description, and an image, 
          and is instructed to output how well the feature applies to the image on a scale from 0 to 4. 
          The LLM output is normalized to a value between 0 and 1 (0.0, 0.25, 0.5, 0.75, 1.0).
          }
  \label{fig:applicability_evaluation}
\end{figure}

The feature exploration phase of the Prediction System training is described in Algorithm~\ref{alg:feature_exploration_phase}, and the model training phase is described in Algorithm~\ref{alg:feature_selection_phase_2}. 
The notation used in these algorithms is defined as follows. 
Specifically, once the linguistic description \(s_{\mathrm{f}}\) of a new feature \(f\) is generated, the LLM automatically evaluates its applicability \(a_{\mathrm{f}}\) for each image. 
An example of applicability evaluation is shown in Fig.~\ref{fig:applicability_evaluation}.

During the feature exploration phase, multiple feature candidates are generated and evaluated; \(F_{\mathrm{accepted}}\) denotes the set of candidates accepted as explanatory variables, and \(F_{\mathrm{rejected}}\) denotes those rejected as unsuitable. 
Here, \(y_{\mathrm{pred}}\) and \(y_{\mathrm{true}}\) denote the predicted score and ground-truth value, respectively, and 
\(e = y_{\mathrm{true}} - y_{\mathrm{pred}}\) denotes the prediction error. 

In the model training phase, \(F_{\mathrm{accepted}}\) undergoes a further screening process, yielding a refined candidate set \(F_{\mathrm{screened}}\). 
Here, \(\ell\) denotes the MAE on \(D_{\mathrm{val}}\), and \(\ell_{\mathrm{min}}\) denotes the minimum MAE observed so far. 
\(m_{\mathrm{best}}\) denotes the model that achieved it. 

In Algorithm~\ref{alg:feature_selection_phase_2}, \(H\) denotes the set of hyperparameter combinations to be explored, with an individual combination denoted \(h\).

Algorithm~\ref{alg:feature_selection_phase_1} is also provided for the ablation study described later. 
It uses forward feature selection, 
where the feature \(f\) from  \(F_{\mathrm{screened}}\) that yields the smallest prediction error when added to \(F_{\mathrm{best}}\) is iteratively selected, with at most one feature per cluster. 
Here, the subset of \(F_{\mathrm{screened}}\) belonging to cluster \(c\) is denoted \(F_{\mathrm{screened},c}\); 
the trial feature set is denoted \(F_{\mathrm{trial}}\); and the set of features accepted as the final explanatory variables is denoted \(F_{\mathrm{best}}\).
At each iteration, \(\ell'_{\mathrm{min}}\) denotes the minimum MAE within the current iteration, \(f'\) the candidate feature to be added, and \(m'\) the provisional best model within the current iteration. 
\(\ell_{\mathrm{thre}}\) denotes the threshold for the improvement in MAE required to continue the forward selection. 

The clustering method used in both the feature exploration phase and the model training phase is described as follows. 
Each feature is represented by a vector of its applicability values across multiple images.
Clustering is performed based on the absolute correlation between these vectors, with the aim of grouping semantically similar features together. 
Specifically, hierarchical clustering is used, where the distance between two feature vectors is defined as \(1 - |\mathrm{corr}|\), and the inter-cluster distance is measured using the average linkage method. 
\(C\) denotes the resulting cluster set.

\begin{algorithm}
\caption{Feature Exploration Phase}
\begin{algorithmic}[1] 
    \State Initialize \(F_{\mathrm{accepted}}\) and \(F_{\mathrm{rejected}}\).
    \For{\(n_{\mathrm{iter}}^\mathrm{in}\) iterations}
        \State Randomly split \(D_{\mathrm{tr}}\) into \(D_{\mathrm{tr}}^{\mathrm{in}}\) and \(D_{\mathrm{val}}^{\mathrm{in}}\).
        \State Cluster the existing feature set \(F_{\mathrm{accepted}}\) using \(D_{\mathrm{tr}}^{\mathrm{in}}\).
        \For{\(n_{\mathrm{model}}\) iterations}
            \State Randomly select up to \(n_{\mathrm{selection}}\) features from \(F_{\mathrm{accepted}}\), with at most one feature per cluster \(c \in C\).
            \State Train a new predictive model \(m\) on \(D_{\mathrm{tr}}^{\mathrm{in}}\) using the selected features. 
            \State Predict on \(D_{\mathrm{val}}^{\mathrm{in}}\) using \(m\). 
            \State Present to an LLM the top \(n_{\mathrm{pos}}\) images with the largest positive prediction errors, along with \(y_{\mathrm{true}}\), \(e\), the feature set used in \(m\) (pairs of \(s_{\mathrm{f}}\) and \(a_{\mathrm{f}}\)), regression coefficients, and the existing sets \(F_{\mathrm{accepted}}\) and \(F_{\mathrm{rejected}}\). Prompt the LLM to devise up to \(n_{\mathrm{candidate}}\) new feature candidates. 
            \State Repeat the above for the top \(n_{\mathrm{neg}}\) images with the largest negative prediction errors. 
        \EndFor
        \State Evaluate the newly devised feature candidates using \(D_{\mathrm{tr}}\). Reject a candidate if its mean applicability across images in \(D_{\mathrm{tr}}\) is below \(a_{\mathrm{thre}}\), or if the absolute correlation with \(y_{\mathrm{true}}\) is below \(r_{\mathrm{thre}}\); otherwise, accept it. Store accepted features in \(F_{\mathrm{accepted}}\) and rejected features in \(F_{\mathrm{rejected}}\).
    \EndFor 
    \State \Return \(F_{\mathrm{accepted}}\) and \(F_{\mathrm{rejected}}\).
\end{algorithmic}
\label{alg:feature_exploration_phase}
\end{algorithm}

\begin{algorithm}
\caption{Model Training Phase (Hyper-Parameter Search) }
\begin{algorithmic}[1] 
    \State Cluster the features in \(F_{\mathrm{accepted}}\) using \(D_{\mathrm{tr}}\). For each cluster \(c \in C\), extract the top \(n_{\mathrm{screened}}\) features by absolute correlation with \(y_{\mathrm{true}}\) and store them in \(F_{\mathrm{screened}}\). 
    \State Initialize \(\ell_{\mathrm{min}} = \infty\) and \(m_{\mathrm{best}} = \emptyset\). 
    \For{\(h \in H\)}
        \State Train a new predictive model \(m\) on \(D_{\mathrm{tr}}\) using the feature set \(F_{\mathrm{screened}}\) with \(h\).
        \State Evaluate \(\ell\) of \(m\) on \(D_{\mathrm{val}}\). 
        \If{\(\ell < \ell_{\mathrm{min}}\)}
            \State Update \(\ell_{\mathrm{min}}  = \ell\) and \(m_{\mathrm{best}} = m\).
        \EndIf 
    \EndFor
    \State \Return \(m_{\mathrm{best}}\).
\end{algorithmic}
\label{alg:feature_selection_phase_2}
\end{algorithm}

\begin{algorithm}
\caption{Model Training Phase (Forward Selection) }
\begin{algorithmic}[1] 
        \State Cluster the features in \(F_{\mathrm{accepted}}\) using \(D_{\mathrm{tr}}\). For each cluster \(c \in C\), extract the top \(n_{\mathrm{screened}}\) features by absolute correlation with \(y_{\mathrm{true}}\) and store them in \(F_{\mathrm{screened}}\). 
        \State Initialize \(F_{\mathrm{best}} = \emptyset\), \(\ell_{\mathrm{min}} = \infty\) and \(m_{\mathrm{best}} = \emptyset\). 
        \For{\(n_{\mathrm{iter}}^\mathrm{out}\) iterations}
            \State Initialize \(\ell'_{\mathrm{min}} = \ell_{\mathrm{min}}\), \(f' = \emptyset\), and \(m' = \emptyset\).
            \For{\(c \in C\)}
                \For{\(f \in F_{\mathrm{screened},c}\)}
                    \State Let \(F_{\mathrm{trial}} = F_{\mathrm{best}} \cup \{f\} \).
                    \State Train a new predictive model \(m\) on \(D_{\mathrm{tr}}\) using \(F_{\mathrm{trial}}\). 
                    \State Evaluate \(\ell\) of \(m\) on \(D_{\mathrm{val}}\). 
                    \If{\(\ell < \ell'_{\mathrm{min}}\)}
                        \State Update \(\ell'_{\mathrm{min}} = \ell\), \(f' = f\), and \(m' = m\). 
                    \EndIf
                \EndFor
            \EndFor
            \If{\(\ell'_{\mathrm{min}} < \ell_{\mathrm{min}} - \ell_{\mathrm{thre}}\)}
                \State Add \(f'\) to \(F_{\mathrm{best}}\), update \(\ell_{\mathrm{min}}  = \ell'_{\mathrm{min}}\) and \(m_{\mathrm{best}} = m'\).
                \State Remove the cluster containing \(f'\) from \(C\).
            \EndIf
            \If{\(\ell'_{\mathrm{min}} \geq \ell_{\mathrm{min}} - \ell_{\mathrm{thre}}\) \textbf{or} \(C = \emptyset\)}
                \State \textbf{break}
            \EndIf
        \EndFor
        \State \Return \(F_{\mathrm{best}}\) and \(m_{\mathrm{best}}\).
\end{algorithmic}
\label{alg:feature_selection_phase_1}
\end{algorithm}

\vspace{3cm}

\subsection{Parameter settings}
\label{App-subsec:parameter_settings}
We describe the detailed parameter settings used in the experiments. 
Note that the settings for the ablation study presented later are also included. 

Regarding the choice of predictive models, a standard ordinary least squares linear regression model was used in the feature exploration phase. 
In the model training phase, the hyperparameter search for the gradient boosting regressor (GBR) was performed following Algorithm~\ref{alg:feature_selection_phase_2}. 
For the ablation study, Algorithm~\ref{alg:feature_selection_phase_2} was also applied with Ridge regression (RR) and random forest regressor (RFR), and Algorithm~\ref{alg:feature_selection_phase_1} was applied with linear regression (LR) and Ridge regression (RR). 
All models were implemented using \texttt{scikit-learn}.

In the feature exploration phase, the following parameters were used: 
the maximum number of features generated per LLM call \(n_{\mathrm{candidate}}=3\), 
the number of images presented to the LLM per call \(n_{\mathrm{pos}} = n_{\mathrm{neg}} = 5\),
the number of temporary predictive models constructed per iteration \(n_{\mathrm{model}}=3\), 
the maximum number of explanatory variables per model \(n_{\mathrm{selection}}=10\), 
and the number of iterations \(n_{\mathrm{iter}}^\mathrm{in}=10\). 
At the beginning of the feature exploration phase when \(F_{\mathrm{accepted}}\) is empty---that is, when no features have yet been accepted---model training is skipped, and images with the largest positive and negative ground-truth values \(y_{\mathrm{true}}\) are presented to the LLM to prompt the generation of new feature candidates. 

In the model training phase, 
the maximum number of features stored in \(F_{\mathrm{screened}}\) per cluster was set to \(n_{\mathrm{screened}}=3\).
For Algorithm~\ref{alg:feature_selection_phase_2}, the hyperparameter search space is shown in Table~\ref{tab:hyper_parameter_search}.
In the table, \(\texttt{max\_depth}=\text{\texttt{None}}\) indicates that no limit is imposed on the tree depth, and \(\texttt{max\_features}=\text{\texttt{"sqrt"}}\) indicates that the number of candidate features at each split is set to the square root of the total number of features. 
For Algorithm~\ref{alg:feature_selection_phase_1}, 
the maximum number of outer iterations was \(n_{\mathrm{iter}}^{\mathrm{out}}=10\).

The thresholds were set as follows: 
mean applicability threshold \(a_{\mathrm{thre}}=0.1\), 
absolute correlation threshold \(r_{\mathrm{thre}}=0.1\), 
and MAE improvement threshold \(\ell_{\mathrm{thre}}=0.001\).
To limit the number of combinations and LLM calls, the maximum number of clusters was set to 20.

\begin{table}[htbp]
\centering
\caption[Hyper-parameter search ranges in feature selection phase]{Hyper-parameter search ranges in model training phase.}
\begin{tabular}{ll}
\hline
\textbf{Model} & \textbf{Search Range} \\
\hline
Gradient Boosting Regression (GBR) &
\begin{tabular}[t]{l}
\texttt{n\_estimators} \(\in \{100, 300, 500\}\) \\
\texttt{learning\_rate}  \(\in \{0.05, 0.1\}\) \\
\texttt{max\_depth} \(\in \{2, 3, 4\}\) \\
\texttt{min\_samples\_leaf} \(\in \{1, 3, 5\}\) \\
\texttt{subsample} \(\in \{0.8, 1.0\}\)
\end{tabular}
\\
\hline
Ridge Regression &
\begin{tabular}[t]{l}
\texttt{alpha} \(\in \{0.5, 1.0, 2.0, 4.0\}\)
\end{tabular}
\\
\hline
Random Forest Regression (RFR) &
\begin{tabular}[t]{l}
\texttt{n\_estimators} \(\in \{200, 500\}\) \\
\texttt{max\_depth} \(\in \{\text{\texttt{None}}, 4, 6, 8\}\) \\
\texttt{min\_samples\_leaf} \(\in \{1, 2, 4\}\) \\
\texttt{max\_features} \(\in \{\text{\texttt{"sqrt"}}, 0.3, 0.5\}\)
\end{tabular}
\\
\hline
\end{tabular}
\label{tab:hyper_parameter_search}
\end{table}

\section{Detailed information of experimental setup}

\subsection{Participant study}
\label{App-subsec:participant_study}
In this work, we conducted a participant study in which each participant underwent multiple behavioral measurements and interviews, in order to collect data on individual aesthetic preferences and to evaluate the effectiveness of the trained system.
The study was conducted under the ethical guidelines of the authors' institution and received ethical approval.
Participants were recruited online. 
Prior to the study, participants were informed of potential risks, including visual discomfort from image viewing. 
It was confirmed that they had no psychiatric conditions or panic symptoms in response to visual stimuli, and that they were able to provide responses in Japanese via keyboard input on a personal computer. 
Informed consent was obtained from all participants, including permission to cite the anonymized data in publications and conference presentations, and to provide the anonymized data to external parties solely upon academic research request. 
The study was designed to ensure that no personally identifiable information was included in any of the questionnaire items. 
Although email addresses were collected via Google Forms solely for the purpose of participant coordination, this information was deleted immediately after data collection and was not used in any analysis. 
A total of 30 participants were recruited (all in their 20s; 25 male, 5 female).
Each participant completed all sessions in approximately 3 hours in total and received  compensation of 6,000 JPY, which exceeds the minimum wage in Japan.

\subsubsection{Overall experimental procedure}
The experiment consisted of two sessions with an interval of more than two weeks:
Exp1 covered collection of individual characteristics and aesthetic rating data, and the interview. 
Exp2 covered prediction of others' aesthetic ratings, subjective comparison among predictors, and evaluation of the overall experience.

In Exp1, eight Google Forms were used.
Form F1 provided an overview of the experiment and collected basic information and informed consent. 
Form F2 administered a questionnaire on individual characteristics. 
Form F3 collected aesthetic rating scores for images, and was prepared separately for each of the five semantic categories. 
After completing Forms F1 through F3, the interview was conducted (note: for one participant, Form F3 for the building semantic category was re-administered after the interview due to missing data).
After the interview, Form F4 collected evaluation data on the interview experience. 

Exp2 consisted of two parts, Part A and Part B. 
In Exp2-A, Form F5 was used to administer the task of predicting the aesthetic ratings of other participants. 
In Exp2-B, three Google Forms were used. 
Form F6 was used to measure within-person variability in aesthetic ratings; participants were asked to re-rate the 45 test images, as they had done in Exp1. 
Form F7 measured two types of evaluations: subjective comparison of predictors, and evaluation of AI-generated images based on their aesthetic preferences. As a supplementary task, participants also evaluated the features extracted by the proposed system. 
Form F8 collected evaluation data on the overall experience of Exp2.

\subsubsection{Data collection of individual aesthetic ratings}

\begin{figure}[htbp]
  \centering
  \includegraphics[width=0.5\linewidth, trim=0 250 2300 0, clip]{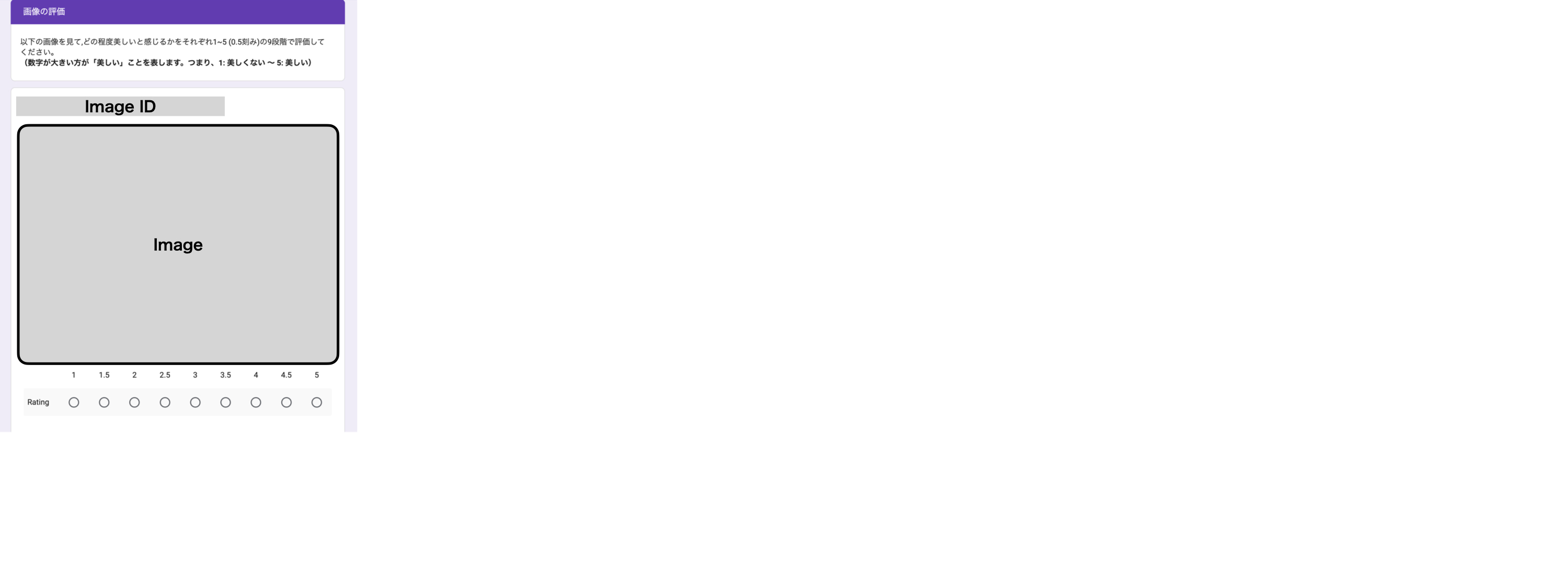}
  \caption[User interface for the interview task]{User interface for the aesthetic rating procedure.}
  \label{fig:rating_ui}
\end{figure}

As described in the main paper, participants rated photographic images randomly sampled from the PARA dataset~\citep{yang2022personalized} on an aesthetic scale from 1.0 to 5.0 (increments of 0.5). 
Five semantic categories were used: portrait, animal, scene, building, and plant.
Images were divided into three rating classes  (Low, Middle, and High) based on mean annotator scores from the dataset, and 20 images were randomly sampled from each category-class combination, yielding \(n_{\mathrm{image}} = 300\) image samples.
The questionnaires were implemented using Google Form and the interface screen observed by participants is shown in Fig.~\ref{fig:rating_ui}.

\subsubsection{Interview procedure}
\begin{figure}[htbp]
  \centering
  \includegraphics[width=\linewidth, trim=0 0 100 0, clip]{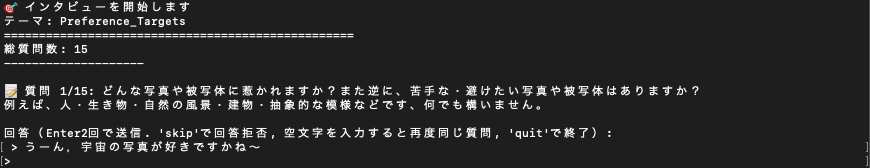}
  \caption[User interface for the interview task]{User interface for the interview phase.}
  \label{fig:interview_ui}
\end{figure}

The interview was conducted in a text-based chat format via a command-line interface. 
The interface screen observed by participants is shown in Fig.~\ref{fig:interview_ui}.
For each participant, the Interview System was launched separately for each of the three interview categories. 
The Preference Targets category consisted of 15 questions, the Image-Evoked Reactions category 10 questions, and the Personal Tastes category 10 questions. 
The Interviewer agent was instructed to cover multiple sub-topics within each category while taking into account the remaining number of questions. 
At the end of each interview, the Interviewer agent generated a summary comment consolidating the target individual's aesthetic tendencies and characteristics based on the accumulated analysis results. 

\newpage
\subsection{Predictor configurations}
\label{App-subsec:predictor_configurations}
The test set of 45 images out of 300 images was held out and shared across predictors.

\subsubsection{The proposed system}
Regarding the LLMs used, both the Interviewer agent and the Analyzer agent employed 
Claude Sonnet 4.5 (claude-sonnet-4-5-20250929)~\citep{anthropic2025introducingclaudesonnet45}.
For the Prediction System, 
Claude Sonnet 4.5 (claude-sonnet-4-5-20250929) was used for feature generation, 
Gemini 2.5 Flash Lite (gemini-2.5-flash-lite)~\citep{google2025gemini25} was used for applicability evaluation, and Gemini 2.5 Flash (gemini-2.5-flash) was used for retry processing upon API call errors. 
In all cases, the temperature was set to 0.0 and the maximum output token length to 4096. 

For the ML module, we adopted a gradient boosting regressor (GBR) as the predictive model, chosen for its ability to flexibly capture nonlinear relationships between features and ratings as well as feature interactions in aesthetics assessment. 

For the DL module, the DL-based predictor described in the following subsection was trained across five independent runs in advance, and the mean of their predicted scores was used as input to the ML module.

\subsubsection{Baseline 1: DL-based predictor}
\begin{figure}[htbp]
  \centering
  \includegraphics[width=0.55\linewidth, trim=0 300 1650 0, clip]{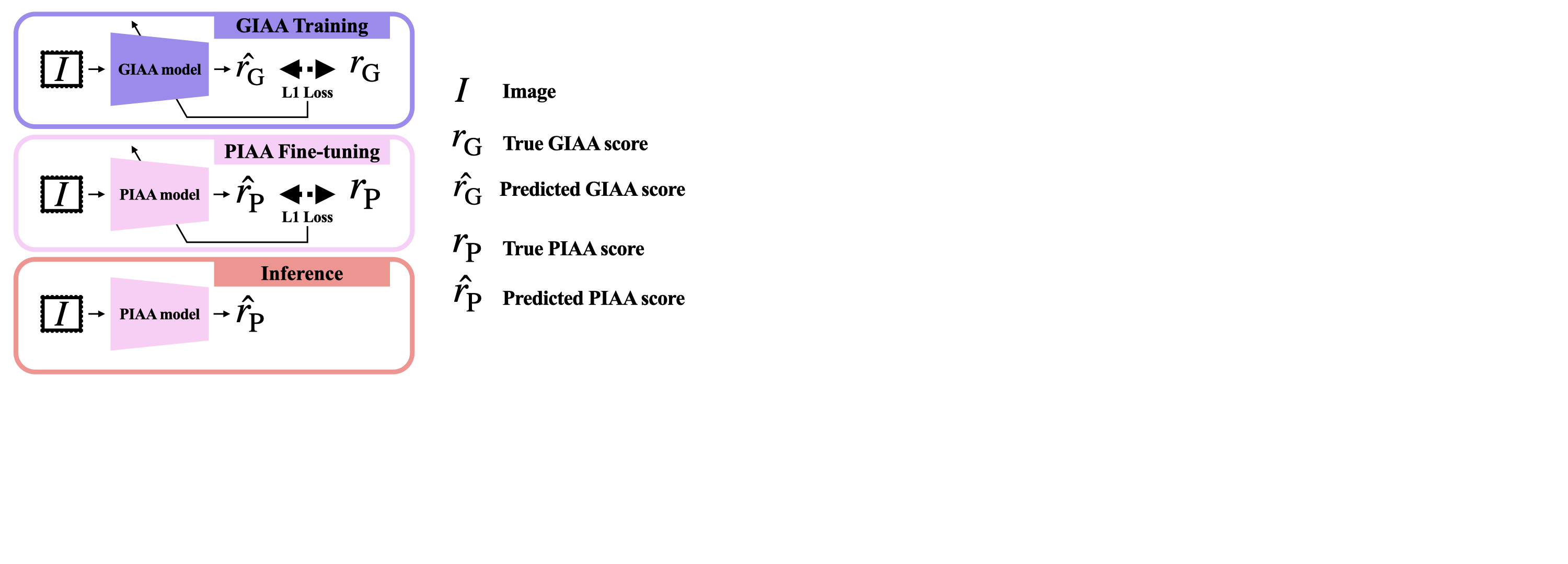}
  \caption[Overview of the DL system]{Overview of the DL-based predictor.}
  \label{fig:dl_system}
\end{figure}

The DL-based predictor (Fig.~\ref{fig:dl_system}) was implemented following the prior study \citep{abe2025quantitative}. 
It was a convolutional neural network with a ResNet-50 backbone~\citep{he2015deep} pretrained on ImageNet. 
The final linear output layer was modified to output a one-dimensional scalar value. 
The model was first trained on the population mean aesthetic evaluation scores from the PARA dataset, 
and then fine-tuned on the target individual's aesthetic evaluation scores. 
An L1 loss function was used with the Adam optimizer (learning rate: 0.001).
During fine-tuning, 45 images were held out as test set images and the remaining 255 images were randomly split into a training set (210 images) and a validation set (45 images). 
To account for training randomness, the final prediction was obtained by averaging the outputs of five independently trained model instances. 

\newpage
\subsubsection{Baseline 2: LLM-based predictor}
\begin{figure}[htbp]
  \centering
  \includegraphics[width=0.55\linewidth, trim=0 350 1880 0, clip]{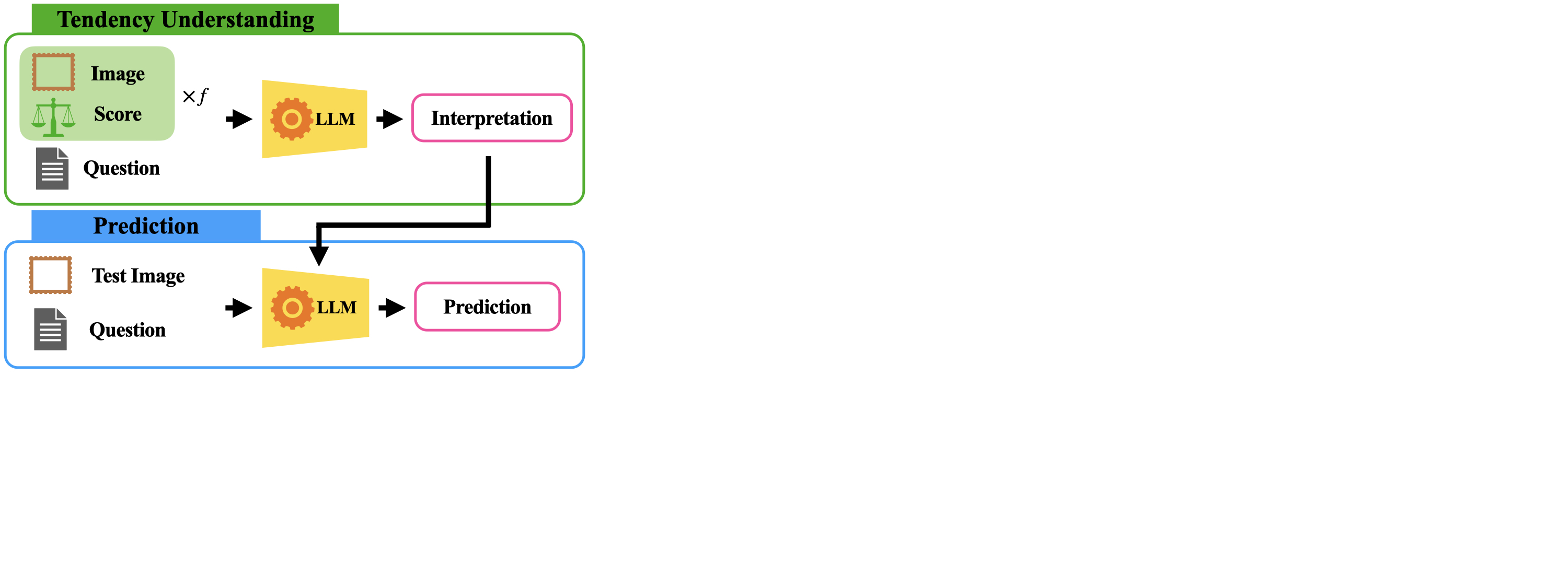}
  \caption[Overview of the LLM system]{Overview of the LLM-based predictor.}
  \label{fig:llm_system}
\end{figure}

The LLM-based predictor (Fig.~\ref{fig:llm_system}) was implemented following the prior study \citep{abe2025harnessingthepowerofllms}. 
Specifically, the system first inferred the target individual's aesthetic tendencies from few-shot examples, and then predicted the aesthetic rating of a given test image based on the inferred tendencies. 
The number of few-shot examples was set to \(f = 12\), randomly sampled from the training or validation set (255 images) within the same semantic category as the test image. 
In the tendency understanding phase, the model was prompted to analyze low-level features (TU\_C1-4 setting as described in the prior study~\citep{abe2025harnessingthepowerofllms}).
The model was instructed to predict aesthetic ratings as discrete values from 1.0 to 5.0 in increments of 0.5.

The LLMs used were Claude Sonnet 4.5 (claude-sonnet-4-5-20250929) and 3.7 (claude-3-7-sonnet-20250219)~\citep{anthropic2025introducingclaudesonnet45, anthropic2025claude37}, 
GPT-5 (gpt-5-2025-08-07) and 4o (gpt-4o-2024-08-06)~\citep{openai2025gpt5, openai2024hello}, and 
Gemini 2.5 Flash (gemini-2.5-flash) and 2.0 Flash (gemini-2.0-flash-001)~\citep{google2025gemini25, google2024next}. 
All models were prompted in English with a temperature of 0.0, and a maximum output token length of 8192, except for GPT-5, for which these parameters were not specified.

\subsubsection{Baseline 3: Human predictor}
Each participant predicted the aesthetic ratings of other participants (referred to as target individuals). 
Specifically, each of the 30 participants predicted the ratings of 5 individuals selected from the remaining 29 individuals. 
Each predictor was anonymously presented with the rating data (image-score pairs) of the target individual for the training and validation sets (255 images), along with the target individual's interview data. 
The presented interview data included the Analyzer agent's analysis summaries and the Interviewer agent's concluding summary comment, covering all three theme categories (Preference Targets, Image-Evoked Reactions, and Personal Tastes).
The predictor was first asked to summarize the target individual's aesthetic tendencies concisely in their own words, and then to provide predicted ratings for the 45 test images on a scale from 1.0 to 5.0 in increments of 0.5. 
No time limit was imposed on this task. 
The questionnaires were implemented using Google Form and participants observed a similar interface screen as shown in Fig.~\ref{fig:rating_ui}.

\section{Additional experiments}

\subsection{Ablation study}
\label{App-subsec:ablation_study}

\subsubsection{Predictive model type and use of DL predictions}

As described in Appendix~\ref{App-subsec:parameter_settings}, 
in the model training phase, the hyperparameter search for the gradient boosting regressor (GBR) was performed following Algorithm~\ref{alg:feature_selection_phase_2}. 
For the ablation study, Algorithm~\ref{alg:feature_selection_phase_2} was also applied with Ridge regression (RR) and random forest regressor (RFR), and Algorithm~\ref{alg:feature_selection_phase_1} was applied with linear regression (LR) and Ridge regression (RR). 
In addition, two variants were considered depending on whether the intermediate predicted scores from the DL module were included as input to the ML module, yielding a total of 10 conditions for the first ablation study. 

The results of comparing the proposed system (Prediction System; PS) and the DL-based predictor (DL) are shown in Table~\ref{tab:comparison_dl_and_ps}.
The Mean and Std columns report the mean and standard deviation across 30 participants. 
Regardless of whether the DL module's output was included as input, PS methods using GBR and RFR outperformed those using LR or RR in terms of mean MAE (Mean column), win rate against DL (Win column), and improvement rate (Imp column). 
Regarding the Win column, only the HPS-GBR-withDL method (hyperparameter search with GBR using DL input) and HPS-RFR-withDL method (hyperparameter search with RFR using DL input) exceeded 50\%. 
Similarly, in terms of the Imp column, only these two methods yielded positive values. 
Wilcoxon signed-rank test (significance level 0.05) were conducted between the DL and 10 PS variants (p-Value, Sig, and Better columns). 
Only HPS-GBR-withDL and HPS-RFR-withDL showed significant improvement over the DL, with improvement rates of 2.3\% and 1.4\%, respectively. 

These results suggest that the inclusion of DL-based intermediate predictions as input is beneficial, indicating that low-level features captured by the DL module remain informative for aesthetics assessment. 
In addition, nonlinear models (GBR and RFR) consistently outperform linear models (LR and RR), suggesting that the relationship between high-level semantic features and aesthetic ratings is inherently nonlinear.

\begin{table}[htbp]
\centering
\caption[Comparison between the DL-based predictor and the Prediction System]{Comparison between the DL-based predictor and the Prediction System. 
"FS" and "HPS" denote the methods used in the model training phase (FS: Forward Selection; HPS: Hyper-Parameter Search).
"LR", "RR", "RFR", and "GBR" denote the types of predictive models employed (LR: Linear Regression; RR: Ridge Regression; RFR: Random Forest Regression; GBR: Gradient Boosting Regression).
The label "withDL" indicates that the Prediction System uses the predicted outputs from the DL module. 
Let \(e_{\mathrm{DL}}\) and \(e_{\mathrm{PS}}\) denote the mean absolute errors obtained with the DL-based predictor and the Prediction System for a participant, respectively. 
The mean and standard deviation of the MAE across 30 participants are reported as Mean and Std.
The proportion of participants for whom \(e_{\mathrm{PS}} < e_{\mathrm{DL}}\) is reported as Win (\%), where ties were excluded.
The relative improvement for each participant was computed as \((e_{\mathrm{DL}} - e_{\mathrm{PS}}) / e_{\mathrm{DL}}\), and its average across participants is shown as Imp (\%).
The differences in the MAE between the DL-based predictor and the Prediction System across its different configurations were evaluated using Wilcoxon signed-rank tests, 
and the resulting p-values were adjusted using the Benjamini-Hochberg procedure for multiple comparisons.
The significance level was set at 0.05.
The adjusted p-values (p-Value), the statistical significance (Sig), and which system was significantly better (Better) are also reported.
"DL", "PS", and "-" denote that the DL-based predictor, the Prediction System, or neither was significantly better. 
Mean and Std values were rounded to three decimal places, Win and Imp values were rounded to one decimal place, and p-Values were reported with three significant digits. 
}
\begin{tabular}{lrrrrrll}
\toprule
Model & Mean & Std & Win (\%) & Imp (\%) & p-Value & Sig & Better \\
\midrule
DL & 0.561 & 0.110 & - & - & - & - & - \\
\cdashline{1-8}[0.4pt/2pt]
FS-LR & 0.669 & 0.114 & 6.7 & -20.6 & 1.3e-07 & True & DL \\
FS-RR & 0.661 & 0.122 & 10.0 & -18.8 & 5.12e-07 & True & DL \\
HPS-RR & 0.626 & 0.116 & 16.7 & -12.4 & 6.93e-05 & True & DL \\
HPS-RFR & 0.607 & 0.125 & 26.7 & -8.7 & 0.00404 & True & DL \\
HPS-GBR & 0.619 & 0.128 & 26.7 & -11.2 & 0.0026 & True & DL \\
FS-LR-withDL & 0.573 & 0.117 & 33.3 & -2.0 & 0.0405 & True & DL \\
FS-RR-withDL & 0.570 & 0.114 & 33.3 & -1.6 & 0.0405 & True & DL \\
HPS-RR-withDL & 0.569 & 0.114 & 36.7 & -1.4 & 0.0803 & False & - \\
HPS-RFR-withDL & 0.553 & 0.108 & 70.0 & 1.4 & 0.0405 & True & PS \\
HPS-GBR-withDL & 0.549 & 0.116 & 70.0 & 2.3 & 0.0328 & True & PS \\
\bottomrule
\end{tabular}
\label{tab:comparison_dl_and_ps}
\end{table}

\subsubsection{Effect of interview data}
The contribution of the LLM-based interview process for eliciting aesthetic preference information was analyzed. 
Specifically, the performance of the Prediction System trained without interview data was evaluated. 
As in the previous section, a total of 10 conditions were evaluated for the second ablation study. 
The results are shown in Table~\ref{tab:comparison_dl_and_ps_interview_ablation_study} and show that without interview data, no advantage over the DL-based predictor was observed. 
This confirms that the LLM-based semi-structured interview process proposed in this work plays a crucial role in improving prediction performance.

\begin{table}[htbp]
\centering
\caption[Comparison between the DL-based predictor and the Prediction System (Interview Ablation Study)]{Comparison between the DL-based predictor and the Prediction System without interview data. 
"FS" and "HPS" denote the methods used in the model training phase (FS: Forward Selection; HPS: Hyper-Parameter Search).
"LR", "RR", "RFR", and "GBR" denote the types of predictive models employed (LR: Linear Regression; RR: Ridge Regression; RFR: Random Forest Regression; GBR: Gradient Boosting Regression).
The label "withDL" indicates that the Prediction System uses the predicted outputs from the DL module. 
Let \(e_{\mathrm{DL}}\) and \(e_{\mathrm{PS}}\) denote the mean absolute errors obtained with the DL-based predictor and the Prediction System for a participant, respectively. 
The mean and standard deviation of the MAE across 30 participants are reported as Mean and Std.
The proportion of participants for whom \(e_{\mathrm{PS}} < e_{\mathrm{DL}}\) is reported as Win (\%), where ties were excluded.
The relative improvement for each participant was computed as \((e_{\mathrm{DL}} - e_{\mathrm{PS}}) / e_{\mathrm{DL}}\), and its average across participants is shown as Imp (\%).
The differences in the MAE between the DL-based predictor and the Prediction System across its different configurations were evaluated using Wilcoxon signed-rank tests, 
and the resulting p-values were adjusted using the Benjamini-Hochberg procedure for multiple comparisons.
The significance level was set at 0.05.
The adjusted p-values (p-Value), the statistical significance (Sig), and which system was significantly better (Better) are also reported.
"DL", "PS", and "-" denote that the DL-based predictor, the Prediction System, or neither was significantly better. 
Mean and Std values were rounded to three decimal places, Win and Imp values were rounded to one decimal place, and p-Values were reported with three significant digits. 
}

\begin{tabular}{lrrrrrll}
\toprule
Model & Mean & Std & Win (\%) & Imp (\%) & p-Value & Sig & Better \\
\midrule
DL & 0.561 & 0.110 & - & - & - & - & - \\
\cdashline{1-8}[0.4pt/2pt]
FS-LR & 0.615 & 0.144 & 26.7 & -9.8 & 0.00505 & True & DL \\
FS-RR & 0.617 & 0.147 & 20.0 & -10.1 & 0.00291 & True & DL \\
HPS-RR & 0.616 & 0.143 & 20.0 & -9.8 & 0.000943 & True & DL \\
HPS-RFR & 0.598 & 0.137 & 33.3 & -6.5 & 0.0201 & True & DL \\
HPS-GBR & 0.616 & 0.138 & 20.0 & -9.9 & 0.000943 & True & DL \\
FS-LR-withDL & 0.569 & 0.109 & 30.0 & -1.5 & 0.0419 & True & DL \\
FS-RR-withDL & 0.569 & 0.110 & 30.0 & -1.4 & 0.0152 & True & DL \\
HPS-RR-withDL & 0.564 & 0.117 & 46.7 & -0.3 & 0.626 & False & - \\
HPS-RFR-withDL & 0.553 & 0.112 & 70.0 & 1.4 & 0.1 & False & - \\
HPS-GBR-withDL & 0.552 & 0.112 & 66.7 & 1.6 & 0.132 & False & - \\
\bottomrule
\end{tabular}
\label{tab:comparison_dl_and_ps_interview_ablation_study}
\end{table}

\subsection{Comparison with the best-performing human predictors}
An additional detailed comparison of all predictors with human predictors was conducted for each of the 30 target individuals. 
Specifically, a win was counted when the MAE of a given predictor was lower than that of the best-performing human predictor (i.e., the one with the lowest MAE among the five human predictors) for a given target individual.
This was repeated across all 30 target individuals to determine for how many the predictor outperformed the best human predictor. 
The results are shown in Table~\ref{tab:better_than_human_min} and show that for 22 out of 30 target individuals, the proposed system (Prediction System; PS) outperformed the best-performing human predictor. 

\begin{table}[htbp]
\centering
\caption[Win counts and win rates against the best human predictor across 30 target persons]{Wins and win rates against the best-performing human predictor across 30 target individuals.}
\begin{tabular}{lrrr}
\toprule
Predictor & Win Count & Win Rate \\
\midrule
PS & 22 &  0.733 \\
DL & 22 &  0.733 \\
Claude Sonnet 3.7 & 17 &  0.567 \\
GPT-4o & 17 &  0.567 \\
GPT-5 & 16 &  0.533 \\
Self Retest & 14 &  0.467 \\
Claude Sonnet 4.5 & 14 &  0.467 \\
Gemini 2.0 Flash & 10 &  0.333 \\
GIAA (w/o self) & 8 &  0.267 \\
Gemini 2.5 Flash & 5 &  0.167 \\
\bottomrule
\end{tabular}
\label{tab:better_than_human_min}
\end{table}

\subsection{Subjective comparison}

As a supplementary experiment, the effectiveness of the predictors was further examined from a subjective perspective, complementary to the objective MAE-based evaluation. 
Specifically, participants were presented with the predictions of multiple predictors trained on their own ratings (the proposed system, DL-based predictor, three LLM-based predictors, and five human predictors).
They were asked to subjectively select which predictor's outputs most closely matched their own aesthetic ratings in Form F7. 
Note that the predictions of three frontier models were used for the LLM-based predictors in this experiment: Claude Sonnet 4.5, Gemini 2.5 Flash, and GPT-5.

Regarding the experiment procedure, nine images were sampled from the 45 test images to form one set.
For each predictor, the nine images were arranged in order of predicted score; the ten predictors were then presented with anonymized labels in a randomized order. 
Participants were asked to select the top three predictors they felt were closest to their own ratings, and this was repeated for five sets. 
Scores of 3, 2, and 1 were assigned to the 1st, 2nd, and 3rd choices, respectively (0 for all others). 
The scores were averaged across 30 participants for each set, and then averaged across the five sets to obtain the mean score for each predictor. 
For human predictors, the mean score was further averaged across the five predictors.

\begin{figure}[htbp]
  \centering
  \includegraphics[width=0.7\linewidth, trim=0 22 0 0, clip]{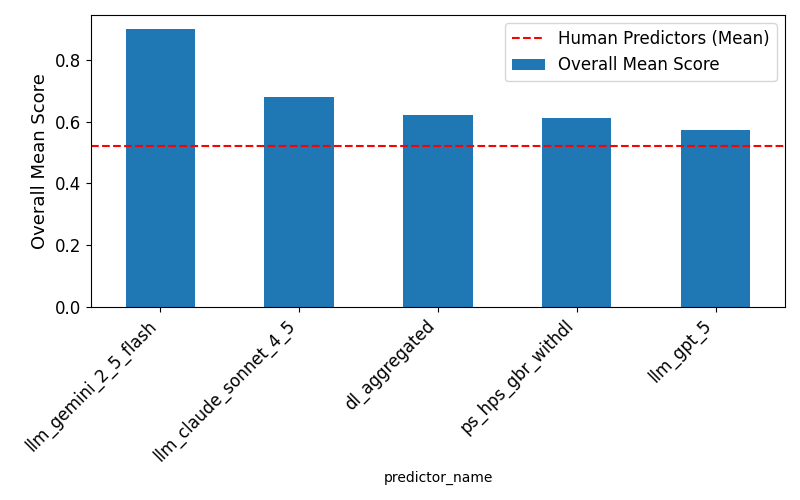}
  \caption[Mean scores of each predictor based on subjective comparisons]{
Mean scores of each predictor based on subjective comparisons.
From left to right, the scores of the LLM-based predictor (Gemini 2.5 Flash), the LLM-based predictor (Claude Sonnet 4.5), the DL-based predictor, the Prediction System, and the LLM-based predictor (GPT-5) are shown. 
The red line indicates the average score of the human predictors.
  }
  \label{fig:subjective_comparison_result}
\end{figure}

The results are shown in Fig.~\ref{fig:subjective_comparison_result}. 
In the subjective evaluation, all AI-based predictors outperformed human predictors.
However, the ranking from best to worst was different from the objective performance measured by MAE in the main paper. 
This suggests that the way humans perceive which predictor is closest to their own aesthetic ratings may differ from the quantitative closeness of predictions over the entire test set. 
One possible explanation is that when humans judge closeness to their own ratings, they may be focusing on a small number of specific cases.

\subsection{Secondary effects of the Interview System}
\label{App-subsec:secondary_effects_of_IS}
After the interview, participants completed Form F4, a questionnaire reflecting on their interview experience. 
The questionnaire items are listed below. 
Responses were collected on a 7-point Likert scale (1: strongly disagree -- 7: strongly agree). 

\begin{enumerate}[
    before=\vspace{0em}, 
    after=\vspace{0em},    
    labelwidth=12em, 
    labelsep=0.5em, 
    align=left,      
    leftmargin=13em  
    ]
    \item [Q1--SelfInsight:] Through this experiment, I gained a clearer understanding of my own aesthetic preferences. 
    \item [Q2--Metacognition:] Through this experiment, I was able to objectively reflect on the reasons behind my aesthetic evaluations.  
    \item [Q3--PerceivedUnderstanding:] I felt that the AI understood my aesthetic preferences well.  
    \item [Q4--SocialPresence:] When interacting with the AI, I felt as if I was communicating with a real presence.  
    \item [Q5--TrustAcceptance:] I felt that the AI's feedback was trustworthy.  
    \item [Q6--PerspectiveExpansion:] Through the interaction with the AI, I began to see images from new perspectives.  
    \item [Q7--AffectiveExperience:] The interaction with the AI was enjoyable and felt comfortable.  
    \item [Q8--ProcessQuality:] The flow of the conversation with the AI felt natural. 
    \item [Q9--AgencyPrivacy:] I felt in control of what information I shared.  
    \item [Q10--FutureUse:] I would like to use AI-based interviews again in the future for self-understanding.  
\end{enumerate}

\begin{figure}[htbp]
  \centering
  \includegraphics[width=\linewidth, trim=0 0 0 0, clip]{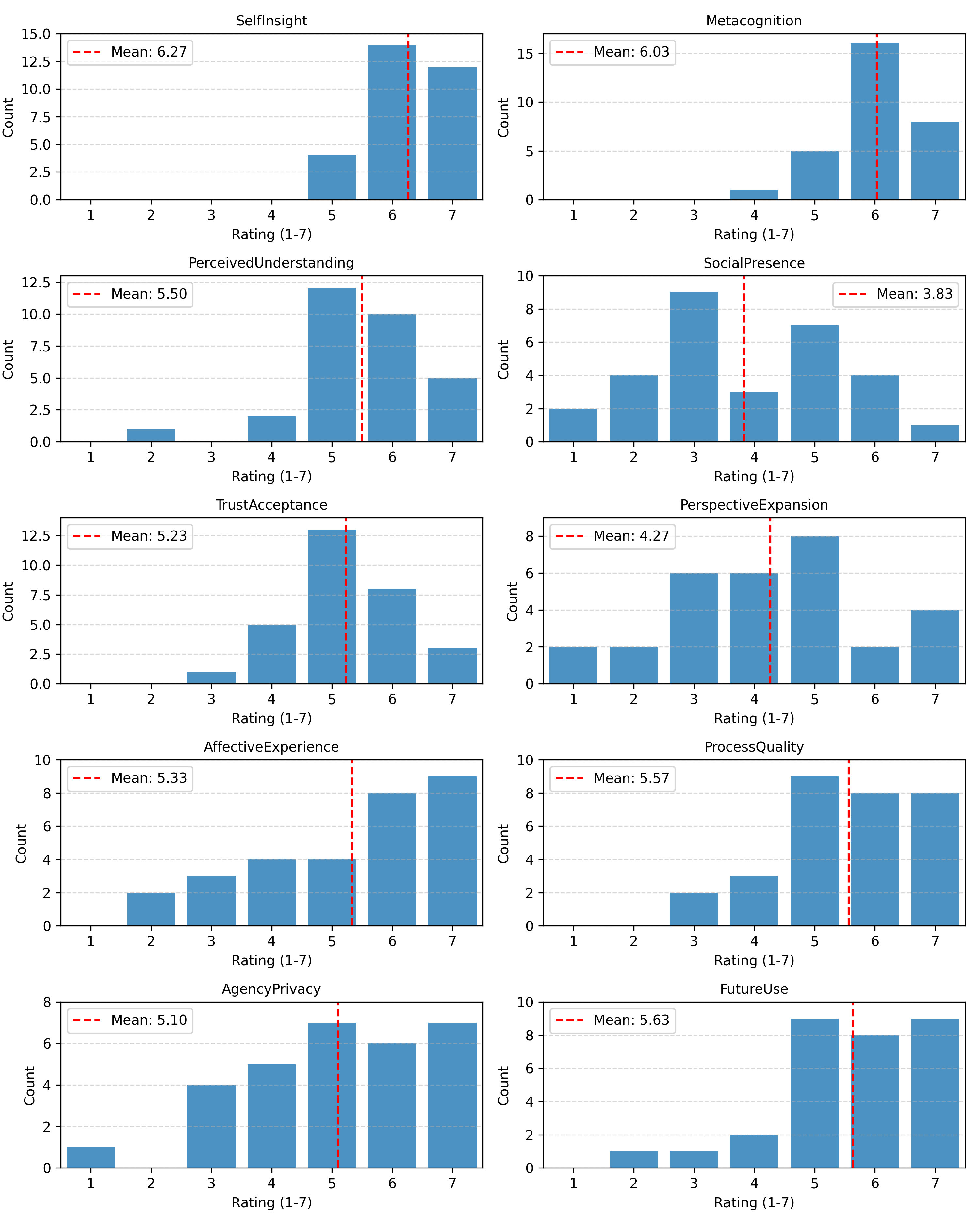}
  \caption[Results of a questionnaire survey on the interview experiences]{Results of a questionnaire survey on the interview experiences. The red dashed lines indicate the mean values.}
  \label{fig:questionnaire_interview}
\end{figure}

The questionnaire results are shown in Fig.~\ref{fig:questionnaire_interview}. The red dashed lines indicate the mean scores.
The SelfInsight and Metacognition items showed scores in the high 6s, considerably above the midpoint of 4 on the Likert scale. 
This indicates that interaction with the Interview System helped participants deepen their self-understanding of their own aesthetic preferences and the reasons behind their evaluations. 
Furthermore, ProcessQuality and FutureUse also scored above 5.5, suggesting that the Interview System offers a high-quality user experience. 

\begin{figure}[htbp]
  \centering
  \includegraphics[width=0.8\linewidth, trim=0 0 0 0, clip]{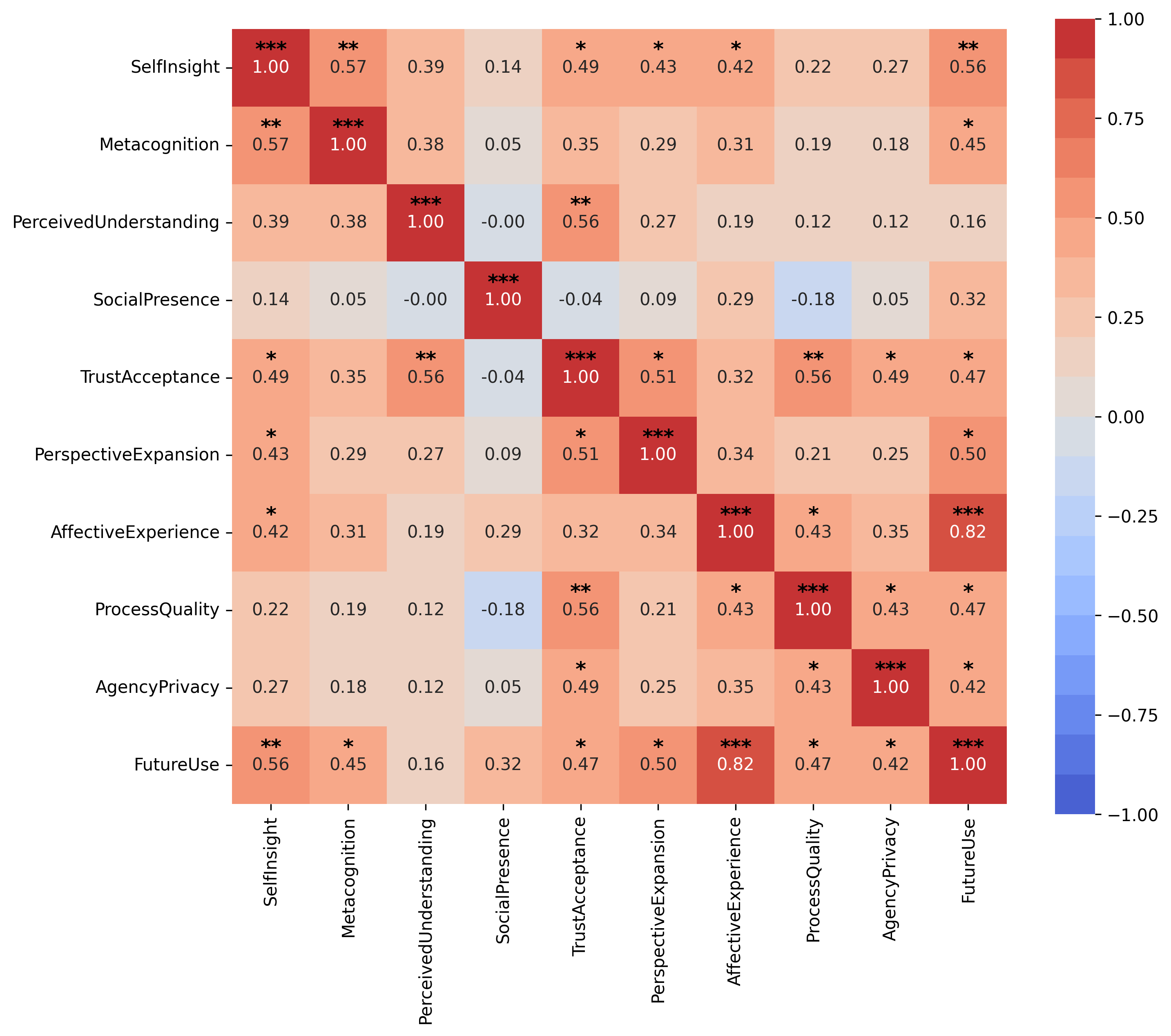}
  \caption[Spearman's rank correlation matrix of evaluation items for the interview experience]{Spearman's rank correlation matrix of evaluation items for the interview experience. 
    To account for multiple comparisons, p-values were adjusted using the Benjamini–Hochberg correction procedure to control the false discovery rate. 
    Asterisks indicate statistical significance after the correction (*: \(p < 0.05\), **: \(p < 0.01\), ***: \(p < 0.001\)). }
  \label{fig:questionnaire_interview_correlation}
\end{figure}

On the other hand, while PerceivedUnderstanding scored 5.50--above the midpoint of 4--SocialPresence scored 3.83, below the midpoint. 
The correlation matrix in Fig.~\ref{fig:questionnaire_interview_correlation} shows that the correlation between these two items was nearly zero.
This indicates that even if participants felt that the AI interviewer understood their aesthetic preferences well, 
they did not necessarily feel that it had a real social presence. 
In other words, the perception of whether an interaction partner understands one's preferences and the perception of whether that partner is socially present may be independently perceived. 

The correlation matrix also reveals a significant strong correlation (0.82) between AffectiveExperience and FutureUse, suggesting that providing an enjoyable and comfortable experience may be important for encouraging users to use the AI system again. 
In contrast, while not statistically significant, a negative correlation (-0.18) was observed between ProcessQuality and SocialPresence, suggesting that higher conversation quality does not necessarily accompany a greater sense of social presence. 
These findings are based on correlational relationships and do not imply causation. 
Further investigation in future research is needed to clarify detailed relationships among these variables.

\section{Broader implications}
\label{App-sec:broader_implications}

The proposed approach of modeling aesthetic value using AI systems has several potential applications and broader societal implications. 

First, by training an AI system on one's own evaluations, individuals can externalize and objectively understand their own aesthetic tendencies. 

Second, training an AI system enables the preservation of one's aesthetic criteria at a given point in time. 
As demonstrated in our experiments, aesthetic evaluations fluctuate over time even within the same individual. 
By preserving aesthetic criteria through AI-based learning and prediction, it becomes possible to reproduce one's past aesthetic evaluations---and even predict evaluations of objects never encountered at that time---simply by loading and running the trained model. 

Third, externalizing one's evaluation criteria through AI enables sharing with others and quantitative comparison of differences. 
Previously, sharing aesthetic criteria between individuals required verbal expression, which is inherently limited, as aesthetic sensibility is often difficult to articulate even for oneself. 
Once an AI system has been trained, however, its behavior can be compared without time constraints. 

Fourth, automating aesthetic evaluation through AI enables individuals to assign their own evaluations to a far greater number of objects than they could encounter in real time. 

In summary, AI-based learning and modeling of aesthetic value holds promise for advancing self-understanding, preservation, sharing, and automation of individual human aesthetic sensibility.

On the other hand, it is also necessary to consider the implications from an ELSI (Ethical, Legal, and Social Implications) perspective. 

The first concern is privacy.
Not only the personal data collected for building the prediction system, but also the trained system itself may serve as a source of personally identifiable information, requiring careful management. 
Both the personal data and the constructed system (including model parameters and prompts) must be appropriately managed to prevent leakage or secondary use. 

The second concern involves moral issues related to human values. 
An aesthetic evaluation prediction system models an individual's evaluation behaviors about what they find good or bad. 
This amounts to externalizing value criteria that have traditionally been inseparable from the individual. 
This may lead to individual-level harms, such as oversimplifying or fixing complex values that fluctuate over time, or promoting distorted self-understanding when personalization accuracy is poor. 
Furthermore, once values can be verbalized and quantified, comparison between value systems becomes easier, potentially leading to societal-level problems such as the loss of diversity in values through selection and elimination. 

Possible approaches to addressing these risks include the following. 
For privacy concerns, identifying the minimal information representation needed to personalize the system, 
and building the system as a local model under individual control, would be effective. 
For the moral challenges related to values, 
it is necessary to continue developing accurate personalization methods for recording, preserving, and learning value systems. 
It is also important to develop methods and interfaces that clearly present the uncertainty of learned values and their divergence from the individual's true underlying values.

\section{Computational resources}
Experiments (model training and evaluation) were conducted across multiple servers. 
The main environments are as follows: 
(1) a server with Intel Xeon Platinum 8268 CPUs (48 cores, 376GiB), NVIDIA GeForce RTX 2080 Ti GPUs (8 units, 11264MiB); 
(2) a server with AMD EPYC 7402 24-Core Processor CPUs (48 cores, 1.0TiB), NVIDIA A100-SXM4-80GB GPUs (4 units, 81920MiB); 
(3) a server with Intel Xeon E5-2697 v4 CPUs (36 cores, 377GiB), NVIDIA RTX A6000 GPUs (2 units, 49140MiB).  

Regarding the computational time and cost, the following are rough estimates based on available logs. 
Training a DL module for the GIAA task took approximately 10 hours per model instance, 
training a DL module for the PIAA task took approximately 10 minutes per individual.
The feature exploration phase for Prediction System, including LLM API calls, took approximately 6 hours per individual and the model training phase took approximately 1 minute per individual.
The estimated total API cost for the feature exploration phase across all 30 target individuals was approximately \$250 USD in total (Claude: \$200, Gemini: \$50).
Additional API costs were incurred for the LLM-based predictor experiments across 30 target individuals: approximately \$300 USD (GPT-5/4o: \$100, Claude Sonnet 4.5/3.7: \$150, Gemini 2.5 Flash/2.0 Flash: \$50).

\section{Licenses for existing assets}
The PARA dataset~\citep{yang2022personalized} was obtained from the official distribution site (\url{https://cv-datasets.institutecv.com/#/data-sets}) and used in accordance with its terms of use, which permit use for academic research purposes.

The ResNet-50 model~\citep{he2015deep} was obtained from the torchvision package (\url{https://docs.pytorch.org/vision/main/models/generated/torchvision.models.resnet50.html}, \url{https://github.com/pytorch/vision/blob/main/LICENSE})  
and is licensed under the BSD 3-Clause License.

\section{Use of LLMs and AI tools}
LLMs are used as core components of the proposed system, 
specifically for conducting semi-structured interviews and extracting semantic features, as described in the main paper. 
In addition, 
LLMs were used during the research process for code assistance and for proofreading, refining, and translating the manuscript. 
All text was originally written by the authors, and all content (texts, codes, and references) was verified by the authors prior to use.

\section{Acknowledgments}
This research was supported by JST SPRING GX project (JPMJSP2108) and JSPS KAKENHI (JP25K24741), Japan. The funding sources had no role in the decision to publish or prepare the manuscript.


\end{document}